\documentclass[12pt]{article}

\usepackage[utf8]{inputenc} % allow utf-8 input
\usepackage[T1]{fontenc}    % use 8-bit T1 fonts
\usepackage{hyperref}       % hyperlinks
\usepackage{url}            % simple URL typesetting
\usepackage{booktabs}       % professional-quality tables
\usepackage{amsfonts}       % blackboard math symbols
\usepackage{nicefrac}       % compact symbols for 1/2, etc.
\usepackage{microtype}      % microtypography
\usepackage{lipsum}
\usepackage{graphicx}
\usepackage{tabularx, array}
\usepackage{makecell}
\usepackage{authblk}

\usepackage{pagecolor,lipsum}

\usepackage{tikz}
\usetikzlibrary{shapes,calc,positioning,arrows}
\usepackage{graphicx}

\usepackage[utf8]{inputenc}
\usepackage{amsmath}
\usepackage{multirow}
\usepackage{xcolor}
\usepackage{epstopdf}
\usepackage{caption}
\usepackage{subcaption}
\usepackage{bm}
\usepackage{pgfplots}
\pgfplotsset{compat=1.15}
\usepackage{verbatim}

\usepackage{footnote}
\usepackage{stfloats}
\usepackage{placeins}
\usepackage{color,soul}
\usepackage{dsfont}
\usepackage{pbox}
\usepackage{enumerate}
\usepackage{etextools}
\usepackage{tabulary}
\usepackage{dsfont}
\usepackage{mathtools}
\usepackage{amssymb}
\usepackage[algoruled, boxed, lined]{algorithm2e}

\tikzstyle{intt}=[draw,text centered,minimum size=6em,text width=5.25cm,text height=0.34cm]
\tikzstyle{intl}=[draw,text centered,minimum size=2em,text width=2.75cm,text height=0.34cm]
\tikzstyle{int}=[draw,minimum size=2.5em,text centered,text width=3.5cm]
\tikzstyle{intg}=[draw,minimum size=3em,text centered,text width=6.cm]
\tikzstyle{sum}=[draw,shape=circle,inner sep=2pt,text centered,node distance=3.5cm]
\tikzstyle{summ}=[drawshape=circle,inner sep=4pt,text centered,node distance=3.cm]

\title{Learned Greedy Method (LGM): A Novel Neural Architecture for Sparse Coding and Beyond}
\author{Rajaei Khatib, Dror Simon and Michael Elad}

\affil{The Computer Science Department - The Technion - Israel}

\date{}

\begin{document}
\maketitle
% \pagecolor{black}
% \color{white}
% \lipsum
%=================================================================
%=================================================================

\begin{abstract}
The fields of signal and image processing have been deeply influenced by the introduction of deep neural networks. These are successfully deployed in a wide range of real-world applications, obtaining state of the art results and surpassing well-known and well-established classical methods. Despite their impressive success, the architectures used in many of these neural networks come with no clear justification. As such, these are usually treated as ``black box'' machines that lack any kind of interpretability. 
A constructive remedy to this drawback is a systematic design of such networks by unfolding well-understood iterative algorithms. A popular representative of this approach is the Iterative Shrinkage-Thresholding Algorithm (ISTA) and its learned version -- LISTA, 
aiming for the sparse representations of the processed signals. In this paper we revisit this sparse coding task and propose an unfolded version of a greedy pursuit algorithm for the same goal. More specifically, we concentrate on the well-known Orthogonal-Matching-Pursuit (OMP) algorithm, and introduce its unfolded and learned version. Key features of our Learned Greedy Method (LGM) are the ability to accommodate a dynamic number of unfolded layers, and a stopping mechanism based on representation error, both adapted to the input. We develop several variants of the proposed LGM architecture and test some of them in various experiments, demonstrating their flexibility and efficiency. 
\end{abstract}
\vspace{0.5in}

%===================================================================================
%===================================================================================

\section{Introduction}
\label{sec:Intro}

In the past decade, Deep Neural Networks (DNN) have been deployed successfully to numerous signal and image processing tasks. This approach has led to state-of-the-art results in various inverse problems, such as image denoising, deblurring, super-resolution, inpainting, and more \cite{DnCNN,FFDNET,DB-NET,SR-NET}, outperforming the more classical model-based and prior-based methods  \cite{K-SVD,BM3D,WNMM,EPLL,RED,Gruop-Sparsity,K-SVD2,Gruop-Sparsity2,GMM-Prior}. In-spite of their remarkable results, most of these DNN architectures lack clear justification and are usually obtained by trial and error. In various cases, this empirical process results in very complex networks with large number (order of millions) of parameters. Consequently, training such a DNN becomes expensive in run-time and memory usage, and a lot of data is required in order to properly learn its parameters. Compared to traditional methods, DNNs are treated as ``black box'' machines, limiting the deployment of these networks in many fields where interpretability is crucial, such as in medical imaging.

An appealing way for alleviating this flaw is by systematically designing networks by unfolding/unrolling iterative algorithms emerging from a prior-based analysis. This line of reasoning stands behind a series of recent publications \cite{Yonina-IEEESPM,Yonina-Deblurring,CSC-MSE,DEEPKSVD,GrishaDenoising,GRUOP_LISTA},
% https://arxiv.org/pdf/1912.10557.pdf - search its IEEE reference
% https://ieeexplore.ieee.org/stamp/stamp.jsp?arnumber=8950351
% https://arxiv.org/pdf/1912.02456.pdf
% search other relevant papers with the same motive
demonstrated to be a viable and attractive alternative to the brute-force practice for designing network architectures. This approach leads to highly interpretable DNN architectures whose structure is well-motivated. An additional benefit to this approach is the fewer number of parameters that it usually requires, easing its learning. The unfolding paradigm dictates various specialized properties on the network, such as parameter sharing between different layers, the non-linearity activation function to use, feedback loops, and more. 

A successful instance of this technique is the LISTA method \cite{LISTA} and its variants \cite{CSC-LISTA,CSC-MSE,GRUOP_LISTA}, in which a fixed number of iterations from the ISTA algorithm \cite{ISTA} is unfolded to a DNN and trained end-to-end. LISTA provides a fast approximation to the ISTA algorithm, aiming to perform sparse coding to the input signals. In this paper we shall focus on this mission of sparse approximation, due to its wide relevance, and offer a DNN alternative to LISTA based on greedy algorithms. 

The field of sparse modeling has gained a lot of interest in the past two decades, both due to its elegant mathematical foundations, and to the applicability of this model to a wide range of data processing tasks \cite{MikiBook}. Intensive work has demonstrated the great relevance of this model as an effective regularizer for inverse problems (for denoising \cite{K-SVD}, deblurring \cite{SparsDebluring}, inpainting \cite{SpaseInpaiting}, demosaicing \cite{SparseDemosaicing}, image-fusion \cite{SparseFusion}, super-resolution \cite{SparseSR}, compressed-sensing \cite{SparseMRICS}, tomographic reconstruction \cite{SparseTR}, MRI imaging \cite{SparseMRICS}, and deraining \cite{Derain2}), as a compression mechanism \cite{K-SVD2,SparseCompression2,SparseCompression3}, and as a feature extractor for recognition tasks \cite{SparseClassification,SparseClassification2,SparseClassification3}. Let us recall the basics of this model, as we rely on these throughout this paper. 

Using an overcomplete dictionary matrix $\bm{D} = \left [ \bm{d}_1 ,\bm{d}_2, \cdots ,\bm{d}_m \right ] \in \mathbb{R}^{n \times m}$ $(m \geq n)$ that contains $m$ atoms ($\bm{d}_k\in \mathbb{R}^{n}$ for $k=1,2,\ldots, m$) as its columns, the sparse-modeling prior assumes that the signal of interest, $\bm{x} \in \mathbb{R}^n$, can be represented as a sparse linear combination of these atoms. Equivalently, this linear combination can be expressed as $\bm{D\alpha}$, where $\bm{\alpha} \in \mathbb{R}^m$ is a sparse vector. This representation may be either exact ($\bm{x} = \bm{D\alpha}$) or approximate ($\left \| \bm{x}-\bm{D\alpha} \right \|_2 \leq \epsilon$). Recovering the sparse representation $\bm{\alpha}$ given $\bm{x}$, $\epsilon$ and $\bm{D}$ requires solving the following NP-hard \cite{NP-HARD} problem:
\begin{eqnarray}\label{eq:P_0^epsilon}
(\textit{P}_{0,\epsilon})~~ \quad \min_{\bm{\alpha}} \|\bm{\alpha}\|_0 ~~~\text{s.t.}~~~  \|\bm{x} - \bm{D\alpha}\|_2 \leq \epsilon, 
\end{eqnarray}
where $\|\bm{\alpha}\|_0$ is the $\ell_0$ ``norm'' that counts the number of non-zero elements in $\bm{\alpha}$. This problem can be approximated using a family of techniques known as \emph{pursuit algorithms}, which can be broadly divided into two categories. The first group of algorithms offers a relaxation of the above optimization, turning it into a continuous problem. For example, $\|\bm{\alpha}\|_0$ can be replaced by $\|\bm{\alpha}\|_1$, forming a convex task known as the Basis Pursuit (BP) objective. BP can be numerically solved using ISTA \cite{ISTA}, which explains the connection to our discussion above. Indeed, running LISTA amounts to learning the appropriate dictionary for the family of signals to be handled, while optimizing the sparse coding performance. More on this will be given in Section~\ref{sec:Lista}.

The second category of pursuit algorithms is the greedy approach that preserves the combinatorial nature of the original problem, recovering the coefficients of $\bm{\alpha}$ sequentially. The Orthogonal Matching Pursuit (OMP) is a popular member of this group \cite{OMP}, and in this work we shall propose a learned unfolded version of it: the $\textit{Learned Greedy method (LGM)}$ network. Similar to LISTA, the parameters of LGM are learned end-to-end through back-propagation. The proposed architecture is characterized by several key and unique features: 
\begin{itemize}
\item This architecture is able of controlling the cardinality of the resulting sparse representation by modifying the number of unfolded layers; 
\item Our scheme can dynamically change the number of layers for each input signal, controlling this way the magnitude of the residual error, akin to the stopping criteria used in OMP \cite{K-SVD,K-SVD2}; and 
\item The resulting network does not utilize the usual element-wise ReLU or shrinkage activation function, but rather imitates the greedy nature of OMP of using the maximal projection thresholding. 

\end{itemize}

\noindent Our work offers several variants of LGM architectures, all inspired by the same origin, while serving different needs:
\begin{itemize}
\item  A simpler greedy method based on the matching pursuit that avoids Least-Squares fitting \ref{subsec:L-MP};
\item  A batch version of the pursuit that aims to speed-up the processing when serving a large group of signals sharing the same dictionary and stopping criterion \ref{subsec:Batch-OMP};
\item  An MMSE-based variant of the OMP (which is a MAP approach), in the spririt of the Random-OMP \cite{RandOMP}; and 
\item A greedy method serving the Convolutional Sparse Coding (CSC) model \cite{CSC1}.
\item A Subspace Pursuit (SP) \ref{subsec:L-SP} compatible algorithm that can operate on groups of non-zeros and remove atoms from the support, empowering it further for handling higher dimensional signals and yielding a more accurate pursuit. 
\end{itemize}

\noindent This work demonstrates the proposed LGM schemes in several experiments. We start with handling of synthetic data, for which our goal is to show its sparse recovery capabilities, and contrast them with LISTA and other pursuit methods. We then move to applications on natural images that operate on image patches, exposing the advantages of LGM over more classical dictionary learning alternatives. We start with an unfolding of the complete K-SVD denoising algorithm for natural images \cite{K-SVD2}, which deploys OMP on the image patches while aiming for global denoising performance. By training this scheme end-to-end, we get competitive denoising performance with LGM as the core engine for the pursuit task. We also demonstrate the LGM on the image deraining application, while relying on a similar architecture to the patch-based denoising, and exploiting a natural division of the dictionary atoms for the separation task. All these tests clearly expose the strength, flexibility, learnability, and usability of the proposed LGM architecture and its variants. 

This paper is organized as follows: Section \ref{sec:Lista} briefly recalls the ISTA method and its learned version LISTA. In Section \ref{sec:Core_LGM} we introduce the core construction of the LGM method. LGM extensions and different variants are discussed in Section \ref{sec:LGM_var}. In Section \ref{sec:syn_experiements} we introduce experimental results for synthetic signals, demonstrating the LGM versus its alternatives. We discuss specific LGM networks for image denoising and deraining in Section \ref{sec:LGM_var2}. Section \ref{sec:Nat_Experiements} presents experimental results of LGM denoising and deraining schemes. We conclude the paper in Section \ref{sec:Conclusion} with a summary of the motivation behind this work and the door it opens for future work. 

%===================================================================================
%===================================================================================

\section{Learned ISTA}
\label{sec:Lista}

%\subsection{ISTA Algorithm}
%\label{subsec:ISTA}
Assume that an ideal signal $\bm{x}^* = \bm{D}\bm{\alpha}^*$ is given to us while being contaminated by bounded energy noise $\bm{x}=\bm{x}^* +\bm{v}$, $\|\bm{v}\|_2 \le \epsilon$. Recovering $\bm{x}^*$ from $\bm{x}$ amounts to solving 
$(\textit{P}_{0,\epsilon})$, as posed in Eq.~$\eqref{eq:P_0^epsilon}$. 
ISTA \cite{ISTA} is an iterative method that aims to solve this problem, by considering the following convex relaxation alternative, in which the $L_0$ is replaced by $L_1$,
\begin{eqnarray}\label{eq:P_1^epsilon}
(\textit{P}_{1,\epsilon})~~\min_{\bm{\alpha}} \|\bm{\alpha}\|_1 ~~~\text{s.t.}~~~  \|\bm{x} - \bm{D\alpha}\|_2 \leq \epsilon.
\end{eqnarray}
This problem can be equivalently written in its unconstrained Lagrangian form,
\begin{equation}\label{eq:Q_1^epsilon}
(\textit{Q}_{1,\lambda})~~\min_{\bm{\alpha}}    \lambda\|\bm{\alpha}\|_1 +\frac{1}{2} \|\bm{x} - \bm{D\alpha}\|_2^2.
\end{equation}
ISTA solves this via the following iterative process:
\begin{eqnarray}\label{eq:ISTA}
\widehat{\bm{\alpha}}_0 & = & \bm{0} \nonumber \\
\widehat{\bm{\alpha}}_t & = & \bm{S}_{\lambda/c} \left( \widehat{\bm{\alpha}}_{t-1} + \frac{1}{c} \bm{D}^T\left ( \bm{x}-\bm{D}\widehat{\bm{\alpha}}_{t-1} \right )   \right ) ~~\mbox{for}~~t=1,2,~\ldots 
\end{eqnarray}
where\footnote{$\lambda_{\text{max}}(\bm{A})$ denotes the largest eigenvalue of $\bm{A}$.} $c > \lambda_{\text{max}}(\bm{D}^T \bm{D})$, and $\bm{S}_{\theta}$ is an element-wise soft-thresholding function defined as 
\begin{eqnarray}
\bm{S}_\theta\left ( v \right ) = \text{sign}\left ( v \right )\max\left ( \left | v \right | -\theta,0 \right ).
\end{eqnarray}
%\begin{itemize}
 % \item $\left [\bm{S}_\theta\left ( \bm{v} \right )  \right ]_i = sign\left ( v_i \right )\left ( \left | v_i \right | -\theta \right )_+$.
  %\item $\bm{S}_\theta\left ( \bm{v} \right )  = \bm{ReLU}\left ( \bm{v}-\theta \right )-\bm{ReLU}\left ( -\bm{v}-\theta \right )$, where $\bm{ReLU}$ is the Rectified Linear Unit function.
%\end{itemize}
ISTA is guaranteed to find the global minimum of the penalty posed in $(\textit{Q}_{1,\lambda})$ in  Eq.~\eqref{eq:Q_1^epsilon}~\cite{ISTA}. This, however, does not imply that we have solved the problem $(\textit{P}_{1,\epsilon})$ in Eq.~\eqref{eq:P_1^epsilon}, as the migration from the choice of $\epsilon$ to $\lambda$ is signal-dependent and non-trivial. This means that solving $(\textit{P}_{1,\epsilon})$ with ISTA should include a search for $\lambda$ so as to satisfy the constraint $\|\bm{x} - \bm{D\alpha}\|_2 \leq \epsilon$. 
Moreover, being able to solve $(\textit{Q}_{1,\lambda})$ (or even $(\textit{P}_{1,\epsilon})$) does not imply that we have necessarily managed to approximate the solution of $(\textit{P}_{0,\epsilon})$ in Eq.~\eqref{eq:P_0^epsilon} -- the original sparse approximation problem we have embarked from. Theoretical guarantees for the proximity between the obtained and the desired solutions do exist (e.g.,~\cite{DonohoEladTemlyakov2006,Tropp-JustRelax,BenHaim2010}), depending on the sparsity of the sought solution and the properties of the dictionary $\bm{D}$. 

\begin{figure}[h]
    \centering
    \begin{subfigure}[b]{0.8\textwidth}
        \includegraphics[width=\textwidth]{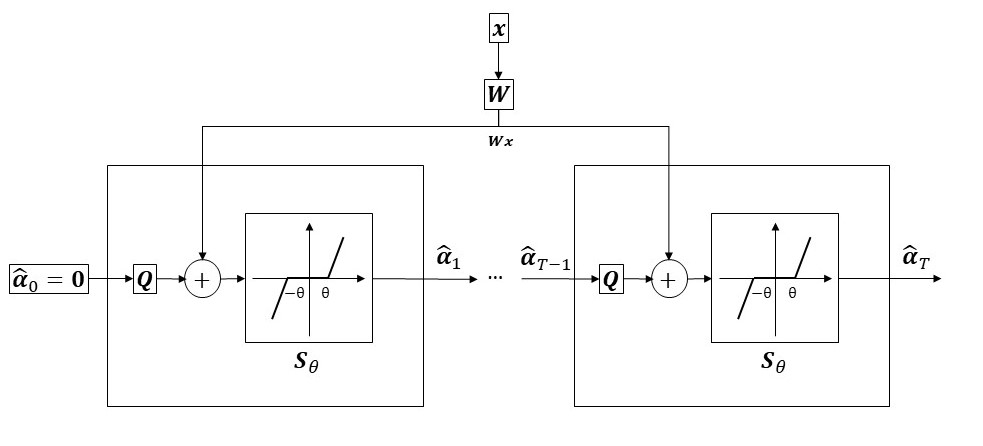}
        \caption{LISTA encoder}
        \label{subfig:Encoder}
    \end{subfigure}
    
    \begin{subfigure}[b]{0.8\textwidth}
        \includegraphics[width=\textwidth]{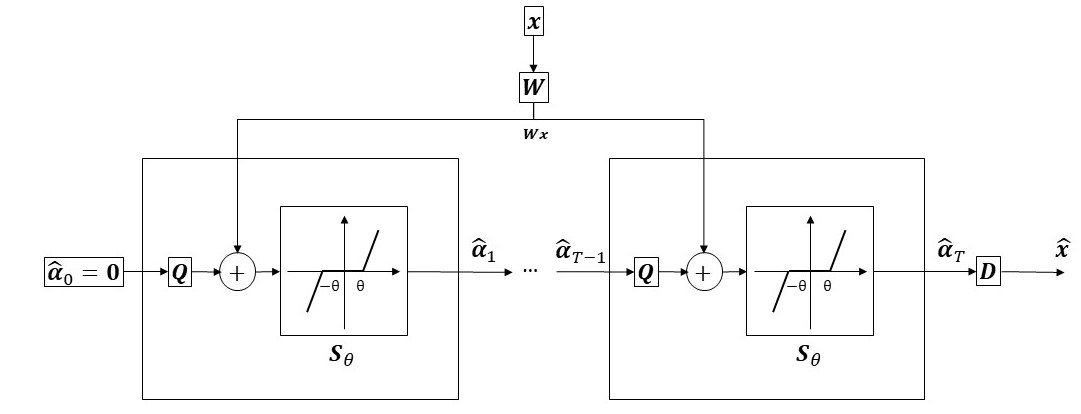}
        \caption{LISTA autoencoder}
        \label{subfig:Autoencoder}
    \end{subfigure}

    \caption{LISTA general architectures}
    \label{fig:LISTA_Arch}
\end{figure}

We move now to the learned variation of ISTA, as originated in~\cite{LISTA}. This starts with an alternative and equivalent formation of the iterative relation in Eq.~\eqref{eq:ISTA} as
\begin{eqnarray}\label{eq:ISTA2}
\widehat{\bm{\alpha}}_t = \bm{S}_{\theta} \left( \bm{Q}\widehat{\bm{\alpha}}_{t-1} +  \bm{W} \bm{x}  \right ),
\end{eqnarray}
where $\bm{Q}=\bm{I}-\frac{1}{c} \bm{D}^T\bm{D} \in \mathbb{R}^{m\times m}$, $\bm{W}=\frac{1}{c}\bm{D}^T \in \mathbb{R}^{m \times n} $ and $\theta=\frac{\lambda}{c}$.
In \cite{LISTA}, a LISTA encoder network is introduced, suggesting a fast sparse coding approximation that comes to solve $(\textit{Q}_{1,\lambda})$. This network, denoted by $\mathcal{F}_T\left ( \bm{x}; \bm{\Theta} \right )$, consists of $T$ unfoldings of the ISTA algorithm, in which each layer follows Eq.~\eqref{eq:ISTA2} in a recurrent manner. Furthermore, the thresholding parameter $\theta$ is extended to a vector of thresholds, allowing for a different treatment for each element. Figure \ref{subfig:Encoder}  describes the LISTA encoder architecture. The parameters of this model, $ \bm{\Theta} = \left ( \bm{Q},\bm{W},\bm{\theta} \right ) $, are to be learned in a supervised fashion. This is done while relying on a sufficiently rich dataset of noisy signals and their true sparse representations, $\left \{ \left ( \bm{x}_i , \bm{\alpha}_i \right ) | \bm{x}_i \in \mathbb{R}^n ,\bm{\alpha}_i \in \mathbb{R}^m  \right \}_{i=1}^{r}$.  The learning amounts to a minimization of the $\ell_2$ loss function 
\begin{eqnarray}
\mathcal{L}_{LISTA}= \frac{1}{r} \sum_{i=1}^{r}{ \left \| \mathcal{F}_T\left ( \bm{x}_i; \Theta \right ) - \bm{\alpha}_i \right \|_2^2},
\end{eqnarray}
obtained through back-propagation and stochastic gradient descent. 
Note that gathering the training set for this learning assumes that $\bm{D}$ is known and available. In addition, we assume that we can solve $(\textit{Q}_{1,\lambda})$ for each of the input signals $\{\bm{x}_i\}_{i=1}^r$ so as to get the appropriate reference representations, $\{\bm{\alpha}_i\}_{i=1}^r$. Note that this requires choosing a specific $\lambda$ to work with. And indeed, under these assumptions, LISTA has been shown to perform favourably, providing a much faster sparse approximation when compared to a direct use of ISTA.
Recent work provides a theoretical analysis of LISTA's speed of convergence and origins of success~\cite{chen2018theoretical, liu2018alista}.

An appealing alternative to the above is an autoencoder mode of learning, where the sparse representation vectors $\{\bm{\alpha}_i\}_{i=1}^r$ are no longer needed in the training set. Keeping the encoding architecture of $\mathcal{F}_T\left ( \bm{x}; \bm{\Theta} \right )$ and adding a decoding layer to its output, $\widehat{\bm{x}} = \bm{D} \widehat{\bm{\alpha}}_T$, we can train this machine end-to-end with a dataset of the form $\left \{ \bm{x}_i , \bm{x}_i^* \right \}_{i=1}^r$, of noisy signals and their clean origins. In this case, the decoder's matrix $\bm{D}$ joins the learned parameters. Figure \ref{subfig:Autoencoder}  describes the LISTA autoencoder architecture. This approach has led to exciting results in various inverse problems, including super-resolution \cite{LISTA_SR}, compressed sensing \cite{LISTA-CS}, image demoisaicking \cite{GRUOP_LISTA}, and image denoising \cite{CSC-MSE,DEEPKSVD}. In all these cases, LISTA yields results that are on par with other DNN models, while having an interpretable architecture that contains much fewer parameters. 

\section{LGM Basic Architecture}
\label{sec:Core_LGM}
\subsection{The OMP Algorithm}
\label{subsec:OMP}

The OMP algorithm \cite{OMP} (see Algorithm \ref{algo:OMP}) greedily solves the
\hyperref[eq:P_0^epsilon]{$(\textit{P}_{0,\epsilon})$} by iteratively increasing the support of the sought sparse solution by one non-zero at a time. Specifically, the algorithm initializes a residual vector $\bm{r}=\bm{x}$ and an empty support set $S=\{\}$. Then, in each of its iterations, the algorithm finds the atom most correlated to the current residual (assuming the atoms are normalized) and adds it to the set $S$. 
Denoting by $\bm{D}_S$ the sub-matrix containing the atoms in $\bm{D}$ that are listed in $S$, and by $\bm{\alpha}_S$ the non-zero portion in $\bm{\alpha}$, we update the representation by the following Least-Squares:
\begin{eqnarray}
\label{eq:LS}
\widehat{\bm{\alpha}}_S = \min_{\bm{z}} \left \| \bm{x}-\bm{D}_S \bm{z} \right \|_2^2 = (\bm{D}_S^T \bm{D}_S)^{-1} \bm{D}_S^T \bm{x}.
\end{eqnarray}
The residual is updated by $\bm{r}=\bm{x}-\bm{D}_S \widehat{\bm{\alpha}}_S$, and the algorithm stops either when this residual is sufficiently small, or when the support reaches a certain cardinality. If the former criterion is used, the number of iterations depends on the given input $\bm{x}$ and the error-threshold. Interestingly, despite its greedy nature, if the original clean signal ($\bm{x}^*$) has a sufficiently sparse representation, OMP is guaranteed to recover a stable solution for the \hyperref[eq:P_0^epsilon]{$(\textit{P}_{0,\epsilon})$} problem \cite{MikiBook}. % For further analysis of the OMP algorithm, the reader is refereed to \cite{MikiBook}.

% is a greedy method that seeks to solve the original sparse recovery problem (\ref{eq:P_0^epsilon}). As described in Algorithm (\ref{algo:OMP}), each iteration of OMP starts by finding the most correlated atom from the dictionary, then adding it to the accumulated support and calculating the new representation by solving a Least-Squares problem. Finally, retrieving the restored signal and its residual. Each iteration of OMP adds a new atom to the representation, e.g. increasing its sparsity by one. The algorithm stops when the residual energy is smaller than a specific threshold or maximum sparsity is archived, this explains why the number of OMP iterations may change from one input to another as mentioned before. 
% OMP algorithm assumes that the dictionary atoms are $L_2$ normalized. An important characteristic of OMP is that it never chooses an atom that has been chosen before, due to the fact that the residual is always orthogonal the chosen atoms. Moreover, the residual is also orthogonal to any linear combination of these atoms, thus any other atom which is a linear combination of them won't be chosed also. This implies that the atoms of $\bm{D}_{S_k}$ are linearly independent, thus the Least-Squares problem has a unique solution . For further analysis of OMP algorithm, the reader is refereed to \cite{MikiBook}. OMP has been deployed in various sparse recovery related applications, such as the K-SVD method \cite{K-SVD2}, demonstrating its flexibility.

\begin{algorithm}[h]
\SetAlgoLined
% \SetKwProg{Input}{Input:}{}{}
\SetKwInOut{Input}{Input}
% \SetKwProg{Output}{Output:}{}{}
\SetKwInOut{Output}{Output}
\SetKwProg{Init}{Init}{}{}
\SetKwFor{For}{for}{do}{endfor}
\SetKw{Break}{break}
\SetKw{Endif}{endif}

\Input{A noisy signal $\bm{x} \in \mathbb{R}^{n}$, a dictionary $\bm{D}\in \mathbb{R}^{n \times m}$, a stopping residual threshold $\epsilon$ and/or a maximum cardinality $s$}
% {
% $\bm{x} \in \mathbb{R}^{n}$: target signal\\
% $\bm{D} =
% \begin{pmatrix}
% | &  & | \\ 
% \bm{d}_1 & ...& \bm{d}_m \\ 
% | &  & |
% \end{pmatrix} 
% \in \mathbb{R}^{n \times m}$: dictionary\\
% $\epsilon \in \mathbb{R}^{+} \cup \left \{ 0 \right \}$: stopping criterion\\
% $s \in \mathbb{N}$: maximal sparsity\\
% }
\Output{A representation vector $\widehat{\bm{\alpha}} \in \mathbb{R}^{m}$, approximating the solution of \hyperref[eq:P_0^epsilon]{$(\textit{P}_{0,\epsilon})$}}
% {
% $\bm{\alpha} \in \mathbb{R}^{m}$: representation under $\bm{D}$
% }

\BlankLine
\Init{
 $\widehat{\bm{\alpha}}_0 = \bm{0}$, $\bm{r}_0 = \bm{x}$, $S_0 = \left \{ \right \}$
}{}
\For{$k=1,2,...,s$}{
$i_0 = \arg\max_i{\left | \bm{d}_i^T \bm{r}_{k-1} \right |} $\\

$S_{k}=S_{k-1} \cup \left \{ i_0 \right \}$\\

$\widehat{\bm{\alpha}}_{S_k} = \left(\bm{D}_{S_k}^T\bm{D}_{S_k}\right)^{-1}\bm{D}_{S_k}^T\bm{x}$ \\

$\widehat{\bm{\alpha}}_k =\begin{cases}\widehat{\bm{\alpha}}_{S_k} & \text{on support} \\
0 &  \text{off support}
\end{cases}$ \\

$\bm{r}_k=\bm{x}-\bm{D}\widehat{\bm{\alpha}}_k$\\

\uIf{$\left \| \bm{r}_k \right \|_2 \leq \epsilon$}{
    \Break \\
  }{\Endif}
}

$\widehat{\bm{\alpha}} = \widehat{\bm{\alpha}}_k$
 
\caption{Orthogonal Matching Pursuit (OMP)}
 \label{algo:OMP}
\end{algorithm}

%=================================================

\subsection{Unfolding OMP}

We now turn to describe our approach of unfolding OMP into a neural network. Generally, each iteration in OMP is transformed into a layer in the proposed architecture. In the original algorithm, the set $S_K$ carries the information from one iteration to the next. Unfortunately, using such a set of indices in a trained network is problematic in terms of differentiability. We  overcome this difficulty by transferring the aggregated dictionary $\bm{D}_{S_k}$ instead. In what follows, we depict the main building blocks of each layer. 

\subsubsection{Maximal-Projection-Thresholding (MPT) Unit}

The $\textit{MPT}$ unit is responsible for deciding which atom will be added to the support in each layer. It operates on the vector $\bm{u}=\bm{D}^T\bm{r}$, representing the correlation of each atom with the current residual. Let $i_0 = \arg\max_i{\left | u_i \right |}$, then the $\textit{MPT}$ function is defined as:
\begin{equation}
\bm{y} = MPT\left ( \bm{u} \right )=
\begin{cases}
u_{i} & i=i_0, \\
0 &\text{otherwise}.
\end{cases}
\label{eq:MPT}
\end{equation}
In other words, when the input is the correlation vector, the output is a vector that contains zeros everywhere except for the index corresponding to the most correlated atom. This function resembles a modified global max-pooling that replaces the regular pooling by a thresholding operator. The gradient of this unit is computed by taking the gradient of the output which propagates back from the next unit, zeroing all its entries except for $i_0$, which remains intact.

%The Jacobian of this function is a matrix that contains zeros everywhere except for the index $(i_0,i_0)$, in which it contains $1$. Consequently, the gradient of this function w.r.t. $\bm{x}$ in the back-propagation stage is calculated by taking the gradient of the next unit, zeroing all its entries except $i_0$ and then passing it to the previous unit. In some way, this function is similar to the max-pooling operation, they both share the functionality of selecting the maximum value (or absolute value) entry.

% An important note to mention, $\textit{MPT}$ function is a semi-continuous function and the gradient calculation discussed here does only hold in the continuous regions (which it is also differentiable)\red{Rajaee: the target of this sentence is to emphasize the dis-continuity of the function, since this concept is rarely used in DNNs. In addition, since we mentioned gradient calculation, some readers will thinks that this function is differentiable (or sub-differentiable)}.
%As will be seen later on this paper, from an empirical point-of-view 

\subsubsection{Atom Selecting (AtoS) Unit}

Following the previous computational step, this unit extracts the selected atom. Given the output of the $\textit{MPT}$ function, $\bm{y} \in \mathbb{R}^m$, and the dictionary $\bm{D}$, this unit yields the atom in $\bm{D}$ corresponding to the index containing the non-zero value in $\bm{y}$.
% In other words, if the only non-zero in $\bm{y}$ resides at index number $i$ then the output of this unit is $\bm{d}_i$ (atom number $i$). 
Equivalently, the $\textit{AtoS}$ unit is defined as follows:
\begin{eqnarray}\label{eq:AS}
\textit{AtoS}\left ( \bm{D},\bm{y} \right ) = \frac{1}{\| \bm{y}\|_{\infty}} \bm{D} \cdot | \bm{y} |.
\end{eqnarray}
The absolute operation is done element-wise, and the $L_\infty$ over $\bm{y}$ simply produces the maximal absolute entry in this vector. Note that the composition of the units $\textit{AtoS}$ and $\textit{MPT}$ provides the functionality of selecting the most correlated atom with the residual $\bm{r}$, i.e., if $\bm{d}_{i_0}$ is indeed this atom, then $\textit{AtoS} \left (\bm{D},\textit{MPT}\left ( \bm{D}^T \bm{r} \right ) \right ) = \bm{d}_{i_0}$. A note worth mentioning, in the description of this unit and the previous one ($\textit{MPT}$), there is a hidden assumption that the atoms of $\bm{D}$ are normalized. This is not the case usually, and later we explain how this assumption is overridden.

\subsubsection{Constructing the LGM Architecture}

\begin{figure}[h]
    \centering
    \includegraphics[width=1\textwidth]{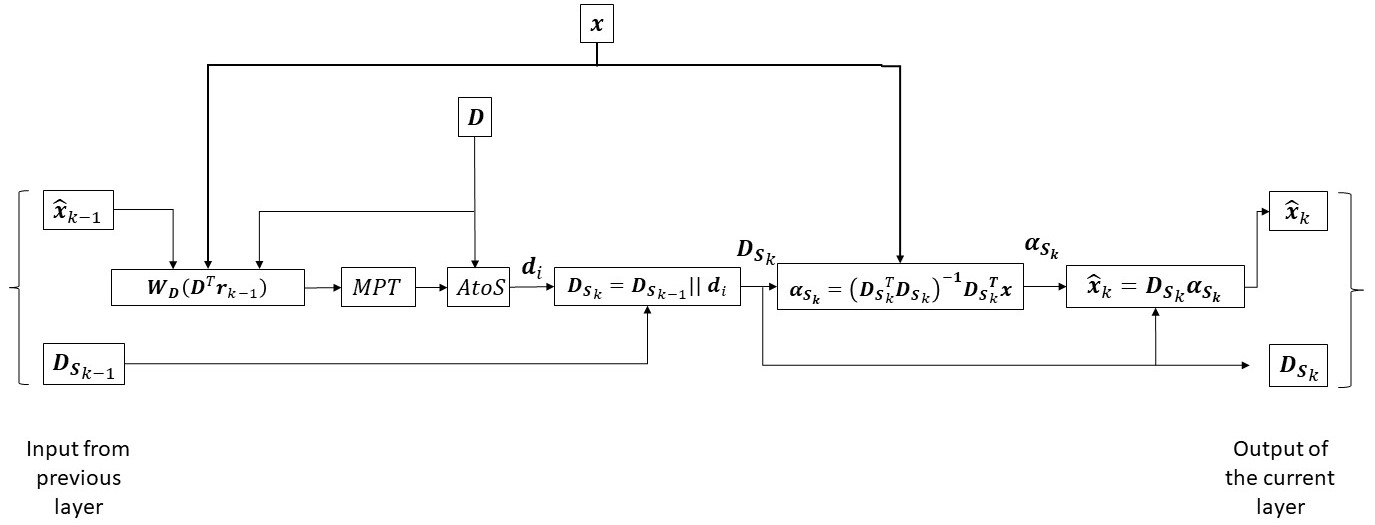}
    \caption{LGM layer}
    \label{fig:OMP_LAYER}
\end{figure}

We turn to describe a single iteration of OMP as a computational graph, which is referred to as an LGM layer (see Algorithm \ref{algo:OMP_LAYER} and Figure \ref{fig:OMP_LAYER}). As mentioned earlier, in the LGM architecture, the aggregated sub-dictionary $\bm{D}_{S_{k}}$ is passed between the different layers instead of the support $S_{k}$. The inputs of each layer are: a signal $\bm{x} \in \mathbb{R}^{n}$, a global dictionary $\bm{D}\in \mathbb{R}^{n \times m}$, the aggregated sub-dictionary from previous layer $\bm{D}_{S_{k-1}}\in \mathbb{R}^{n \times \left ( k-1 \right )}$, and $\widehat{\bm{x}}_{k-1} \in \mathbb{R}^{n}$ which is the restored signal using the atoms of $\bm{D}_{S_{k-1}}$. The LGM layer starts by identifying and adding the most correlated atom with the current residual to the support, by composing the $\textit{MPT}$ and $\textit{AtoS}$ units.
% then passing it to the $\textit{MPT}$ unit, the output of $\textit{MPT}$ is passed to the $AS$ unit alongside with the dictionary $\bm{D}$, and finally the output of $\textit{AS}$ is the new atom to add $\bm{d}_i$ which is the most correlated atom with the residual. 
Then, by using the updated support atoms $\bm{D}_{S_{k}}$, the representation under these atoms $\widehat{\bm{\alpha}}_{S_{k}}$ is computed by solving the corresponding LS problem. Finally, the restored signal $\widehat{\bm{x}}_{k}=\bm{D}_{S_k}\widehat{\bm{\alpha}}_{S_k}$ is calculated. Note that the LS solver ($\widehat{\bm{\alpha}}_{S_k}= \left(\bm{D}_{S_k}^T\bm{D}_{S_k}\right)^{-1}\bm{D}_{S_k}^T\bm{x}$) is a part of LGM layer's computational graph, and throughout back-propagation, the derivatives of the inverse matrix are computed as in \cite{Petersen2008}.

Describing LGM layer as a computational graph is essential in order to build the LGM network (which we are going to introduce next) as a computational graph, and thus it can be trained through back-propagation. However, in the evaluation part, where we just want to apply the model without calculating the derivatives, the model inference can be accelerated by immediately selecting $\bm{d}_{i_0}$ as $i_0$ column in $\bm{D}$ (as in the original OMP algorithm), thus the matrix multiplication operation in $\textit{AtoS}$ is spared.

% A note worth mentioning here, in the evaluation part where we just want to apply the model without calculating the derivatives, the model inference can be accelerated by immediately selecting $\bm{d}_i$ from $\bm{D}$ (similarly to the original OMP algorithm) instead of using $\textit{MPT}$ and $\textit{AtoS}$ as mentioned before.

% The next stage is to create $\bm{D}_{S_{k}}$ by adding $\bm{d}_i$ as a new atom to $\bm{D}_{S_{k-1}}$. The following step is to find the representation vector $\bm{\alpha}_{S_{k}} \in \mathbb{R}^k$ by solving the LS problem (\ref{eq:LS}). The last step is to reconstruct back the restored signal of this layer, and the outputs of this layer are ($\bm{D}_{S_{k}}$ , $\widehat{\bm{x}}_{k}$). 
%An important note to mention here, we could have the same functionality without passing $\widehat{\bm{x}}_{k}$ from each layer to the next one, the reason why we decided to keep it although is to make the architectures simpler and more efficient computationally.
\begin{algorithm}[h]
\SetAlgoLined
\SetKwFunction{FName}{LGMLayer}
\SetKwProg{Pn}{Function}{:}{}

  \Pn{\FName{$\bm{x} \in \mathbb{R}^{n}$, $\bm{D}\in \mathbb{R}^{n \times m}$, $\bm{D}_{S_{k-1}}\in \mathbb{R}^{n \times \left ( k-1 \right )}$, $\widehat{\bm{x}}_{k-1} \in \mathbb{R}^{n}$}}{
  \BlankLine
  $\bm{r}_{k-1} = \bm{x}-\widehat{\bm{x}}_{k-1}$ \\
%   $\bm{u}=\bm{D}^T\bm{r}_{k-1}$
  $\bm{u}=\bm{W}_{\bm{D}}  \bm{D}^T\bm{r}_{k-1}$
  
$\bm{d}_{i_0} =\textit{AtoS} \left (\bm{D},\textit{MPT}\left (\bm{u}  \right ) \right )$    

$ \bm{D}_{S_{k}} = \begin{bmatrix}
\bm{D}_{S_{k-1}}, & \bm{d}_{i_0}
\end{bmatrix}~~~$

% $\bm{D}_{S_{k}}^T \bm{D}_{S_{k}} \bm{\alpha}_{S_{k}} = \bm{D}_{S_{k}}^T \bm{x}$ 
$\widehat{\bm{\alpha}}_{S_k}= \left(\bm{D}_{S_k}^T\bm{D}_{S_k}\right)^{-1}\bm{D}_{S_k}^T\bm{x}$

$\widehat{\bm{x}}_{k} = \bm{D}_{S_{k}} \widehat{\bm{\alpha}}_{S_{k}}$
\BlankLine
  }
  {\KwRet  $\{\bm{D}_{S_{k}}$ , $\widehat{\bm{x}}_{k}\}$}
 
 \caption{LGM Layer Inference}
 \label{algo:OMP_LAYER}
\end{algorithm}

The LGM network is defined in an iterative manner in Algorithm \ref{algo:LOMP_ARCH}. In our architecture, the number of LGM layers used in the network changes w.r.t. each input,  akin to the OMP algorithm. Specifically, we incorporate a sparsity constraint in our network, i.e. maximum number of non-zeros (and thus layers), denoted by $s$, and a residual threshold $\epsilon$ constraint. The parameters of this network are $ \bm{\Theta} = \left ( \bm{D},s,\epsilon \right ) $, where $\bm{D}$ is learned through back-propagation, and $\left (s,\epsilon\right)$ are specified in advance, or manually tuned in case they are not known. However, since $\bm{D}$ is learned, its atoms are not guaranteed to be normalized; thus $\bm{D}^T\bm{r}_{k-1}$ is multiplied by $\bm{W}_{\bm{D}} = diag^{-1} \left ( \left \| \bm{d}_1  \right \|_2, \left \| \bm{d}_2  \right \|_2, ... , \left \| \bm{d}_m  \right \|_2 \right )$ (see Algorithm \ref{algo:OMP_LAYER}).

In many tasks, such as denoising, it is preferred to learn a different dictionary for the operation $\widehat{\bm{x}}_{k} = \bm{D}_{S_{k}} \widehat{\bm{\alpha}}_{S_{k}}$ in the last layer of LGM -- this is the $\textit{Synthesis Dictionary}$. Achieving this requires to slightly change the proposed architecture, and instead of having one dictionary, the network contains two, one as the regular $\bm{D}$ and one as the synthesis dictionary $\bm{D}_2$. More precisely, two sub-dictionaries are passed between the different LGM layers, one is a sub-dictionary of $\bm{D}$ and the other is a corresponding sub-dictionary of $\bm{D}_2$. In each layer, and after selecting $\bm{d}_{i_0}$, the atom with the same index ($i_0$) is also selected from $\bm{D}_2$ in a similar manner by $\textit{AtoS} \left (\bm{D}_2,\textit{MPT}\left (\bm{u}  \right ) \right )$. Then, this atom is added to $\bm{D}_2$ sub-dictionary. Finally, the restored signal is synthesised using the sub-dictionary of $\bm{D}_2$ in the last layer. $\bm{D}$ and $\bm{D}_2$ are initialized equally and learned through back-propagation. From here after, when we reference LGM, it implies two dictionaries unless said otherwise.

% In each layer, when an atom from $\bm{D}$ is selected to be added to the corresponding sub-dictionary, the atom with the same index from $\bm{D}_2$ is selected and added to an additional aggregated sub dictionary which is used only in the last layer in order to create the final restored signal.

%, $s$ and $\epsilon$ must be tuned in advance.

%parameters
%normlize
%output
%back-prob problem
%which is a parameter of the network 

\begin{algorithm}[h]
% \color{white}

\SetAlgoLined
\SetKwProg{Input}{Input:}{}{}
\SetKwProg{Output}{Output:}{}{}
\SetKwProg{Init}{Init}{}{}
\SetKwFor{For}{for}{do}{endfor}
\SetKw{Break}{break}
\SetKw{Endif}{endif}
\SetKwFunction{FName}{LGMLayer}

\Input{$\bm{x} \in \mathbb{R}^{n}$}{}
\Output{$\widehat{\bm{x}} \in \mathbb{R}^{n}$}{}

\Init{
 $\bm{D}_{S_{0}} =\bm{ \left [ ~ \right ] }\in \mathbb{R}^{n \times 0}$, $\widehat{\bm{x}}_{0} = \bm{x}$}{}
\BlankLine
\For{$k=1,2,...,s$}{
$\{ \bm{D}_{S_{k}}~,~\widehat{\bm{x}}_{k}\}$  = \FName{$\bm{x}$, $\bm{D}$, $\bm{D}_{S_{k-1}}$, $\widehat{\bm{x}}_{k-1}$} \\
$\bm{r}_k = \bm{x}-\widehat{\bm{x}}_{k}$ \\
\uIf{$\left \| \bm{r}_k \right \|_2 \leq \epsilon$}{
    \Break \\
  }{\Endif}
}
$\widehat{\bm{x}} = \widehat{\bm{x}}_{k}$

 \caption{LGM Network Inference}
 \label{algo:LOMP_ARCH}
\end{algorithm}

%===================================================================================
%===================================================================================

%Inference
%denoising

\section{LGM Variations}
\label{sec:LGM_var}
In this section we present several LGM variants by unfolding Matching Pursuit (MP) \cite{Mallat1993MatchingPW}, Subspace Pursuit (SP) \cite{SubSpace} and Rand-OMP \cite{RandOMP} greedy methods. We also revisit the Batch-OMP \cite{rubinstein2008efficient} algorithm, which accelerates the inference run-time performance of LGM. In addition, we consider the Convolutional Sparse Coding (CSC) model \ref{subsec:CSC}, and propose a LGM method suited for it, being an unfolded version of the Global Convolutional Matching Pursuit (GCMP) \cite{CSC-GCMP}. We note, however, that we do not provide experimental results for these algorithms, as our focus remains the plain LGM method.

\subsection{Learned-MP: Matching Pursuit Based LGM}
\label{subsec:L-MP}
The MP algorithm \cite{Mallat1993MatchingPW} is a simplified greedy method that seeks to solve the \hyperref[eq:P_0^epsilon]{$(\textit{P}_{0,\epsilon})$} problem. MP is very similar to the OMP, but replaced the Least-Squares update of the coefficients by a simpler computation. After finding the most correlated atom with the current residual $\bm{r}$, MP adds its correlation's coefficient to the corresponding entry in the representation vector. Since we assume no normalized atoms, the coefficient update should take this into account and use $\bm{d}_i^T \bm{r}/\left \| \bm{d}_i  \right \|_2$. Unlike OMP, MP might choose the same atom more than one time, thus the cardinality of $\widehat{\bm{\alpha}}_s$ is always less or equal to the number of iterations - which is $s$. For further analysis and stability guarantees of the MP algorithm, the reader is referred to \cite{MikiBook}. MP is unfolded using the $\textit{MPT}$ unit defined earlier, which given the correlation vector, its output is a vector that contains the coefficient of the most correlated atom in its corresponding entry and zero elsewhere. The Least-Squares step in the OMP is omitted, simplifying the overall network. L-MP inference is described in Algorithm \ref{algo:LMP_ARCH}. Similarly to the process described earlier, this scheme can be extended to include two dictionaries, regular and synthesis.

\begin{algorithm}[h]

\SetAlgoLined
\SetKwProg{Input}{Input:}{}{}
\SetKwProg{Output}{Output:}{}{}
\SetKwProg{Init}{Init}{}{}
\SetKwFor{For}{for}{do}{endfor}
\SetKw{Break}{break}
\SetKw{Endif}{endif}
\SetKwFunction{FName}{LGMLayer}

\Input{$\bm{x} \in \mathbb{R}^{n}$}{}
\Output{$\widehat{\bm{x}} \in \mathbb{R}^{n}$}{}

\Init{
  $\bm{r}_{0} = \bm{x}$ , $\widehat{\bm{\alpha}}_{0} = \bm{0} \in \mathbb{R}^m$}{}
\BlankLine
\For{$k=1,2,...,s$}{
$\widehat{\bm{\alpha}}_k =\widehat{\bm{\alpha}}_{k-1} +  \bm{W}_{\bm{D}}\textit{MPT}\left ( \bm{W}_{\bm{D}} \left ( \bm{D}^T\bm{r}_{k-1} \right ) \right) $ \\
$\widehat{\bm{x}}_{k} = \bm{D}\widehat{\bm{\alpha}}_k$ \\
$\bm{r}_k = \bm{x}-\widehat{\bm{x}}_{k}$ \\
\uIf{$\left \| \bm{r}_k \right \|_2 \leq \epsilon$}{
    \Break \\
  }{\Endif}
}
$\widehat{\bm{x}} = \widehat{\bm{x}}_{k}$

 \caption{L-MP Network Inference}
 \label{algo:LMP_ARCH}
\end{algorithm}

\subsection{Learned-SP: Subspace Pursuit Based LGM}
\label{subsec:L-SP}
The SP algorithm \cite{SubSpace} is a more sophisticated method for solving the \hyperref[eq:P_0^epsilon]{$(\textit{P}_{0,\epsilon})$} problem. The sought sparsity $s$ must be provided in advance, and the cardinality of its recovered representation is exactly $s$ (unlike OMP/MP). SP starts by fining the $s$ most correlated atoms with the signal $\bm{x}$ and adds them to the initial support set $S_0$. Then, the initial residual is calculated by solving the corresponding LS problem, i.e. $\bm{r}_0=\bm{x}-\bm{D}_{S_0}\left(\bm{D}_{S_0}^T\bm{D}_{S_0}\right)^{-1}\bm{D}_{0_k}^T\bm{x}=\bm{x}-\bm{D}_{S_0}\bm{D}_{S_0}^{\dagger}\bm{x}$. Afterwards, in each iteration the $s$ most correlated atoms with the residual are added to a temporarily support set $\widetilde{S}_k$ alongside the atoms from the previous support. Using the $2s$ atoms of $\widetilde{S}_k$, a temporal representation $\widetilde{\bm{\alpha}}_k$ is obtained by solving the corresponding LS problem. The $s$ atoms that make it to the next iteration (i.e. the $S_k$ set) are those with the $s$ largest magnitudes in $\widetilde{\bm{\alpha}}_k$. Next, $\widehat{\bm{\alpha}}_k$ is updated by solving the corresponding LS problem again. The algorithm stops when the residual energy stops decaying. The SP algorithm is described in Algorithm \ref{algo:SubSpace}. For further details about SP and its theoretical stability guarantees, the reader is referred to \cite{SubSpace}. 

SP is unfolded in a similar approach to what was done in order to unfold OMP. Generally, each iteration in SP is transformed into a layer of the proposed architecture. Analogous to the LGM architecture presented earlier, the aggregated sub-dictionary $\bm{D}_{S_k}$ is passed between the different layers of L-SP instead of the chosen atoms support $S_k$. In what follows, we depict the main building blocks of each layer, then we combine them together in order to obtain the L-SP architecture.

\begin{algorithm}[!htbp]
\SetAlgoLined
% \SetKwProg{Input}{Input:}{}{}
\SetKwInOut{Input}{Input}
% \SetKwProg{Output}{Output:}{}{}
\SetKwInOut{Output}{Output}
\SetKwProg{Init}{Init}{}{}
\SetKwFor{For}{for}{do}{endfor}
\SetKw{Break}{break}
\SetKw{Endif}{endif}

\Input{A noisy signal $\bm{x} \in \mathbb{R}^{n}$, a dictionary $\bm{D}\in \mathbb{R}^{n \times m}$, cardinality $s$}

\Output{A representation vector $\widehat{\bm{\alpha}} \in \mathbb{R}^{m}$, approximating the solution of \hyperref[eq:P_0^epsilon]{$(\textit{P}_{0,\epsilon})$}}
% {
% $\bm{\alpha} \in \mathbb{R}^{m}$: representation under $\bm{D}$
% }

\BlankLine
\Init{
 $S_0 = \left \{ s \textup{ indices corresponding to the largest magnitude entries of } \bm{D}^T\bm{x} \right \}$, $\widehat{\bm{\alpha}}_0 =\begin{cases}
\left(\bm{D}_{S_0}^T\bm{D}_{S_0}\right)^{-1}\bm{D}_{S_0}^T\bm{x} & \textit{on support} \\
0 &  \textit{off support}
\end{cases}$, \\
$\bm{r}_0 = \bm{x}-\bm{D}\widehat{\bm{\alpha}}_0$
}{}
\For{$k=1,2,...$}{
$\widetilde{S}_k=S_{k-1} \cup \left \{ s \textup{ indices corresponding to the largest magnitude entries of } \bm{D}^T\bm{r}_{k-1} \right \}$\\

$\widetilde{\bm{\alpha}}_k =\begin{cases}
\left(\bm{D}_{\widetilde{S}_k}^T\bm{D}_{\widetilde{S}_k}\right)^{-1}\bm{D}_{\widetilde{S}_k}^T\bm{x} & \textit{on support} \\
0 &  \textit{off support}
\end{cases}$ \\

$S_k= \left \{ s \textup{ indices corresponding to the largest magnitude entries of } \widetilde{\bm{\alpha}}_k \right \}$\\

$\widehat{\bm{\alpha}}_k =\begin{cases}
\left(\bm{D}_{S_k}^T\bm{D}_{S_k}\right)^{-1}\bm{D}_{S_k}^T\bm{x} & \textit{on support} \\
0 &  \textit{off support}
\end{cases}$ \\

$\bm{r}_k=\bm{x}-\bm{D}\widehat{\bm{\alpha}}_k$\\

\uIf{$\left \| \bm{r}_k \right \|_2 > \left \| \bm{r}_{k-1} \right \|_2$}{
    $\widehat{\bm{\alpha}}=\widehat{\bm{\alpha}}_{k-1}$ \\
    \Break \\
  }{\Endif}
}

\caption{Subspace Pursuit (SP) \ref{subsec:L-SP}}
 \label{algo:SubSpace}
\end{algorithm}

\subsubsection{Maximal-S-Projection-Thresholding (MSPT) Unit}

The $\textit{MSPT}$ unit is responsible for deciding which $s$ atoms will be added to the support in each layer. Given a vector $\bm{u}$, let $I=\left \{ i_1,i_2,\cdots ,i_s \right \}$ denote the indices corresponding to the largest entries of $\left | \bm{u} \right |$. The $\textit{MSPT}$ function is defined as
\begin{equation}
\bm{Y} = MSPT\left ( \bm{u} \right )= \left [ \bm{y}_1,\bm{y}_2,\cdots,\bm{y}_s \right ] \in \mathbb{R}^{m \times s},
\label{eq:MSPT}
\end{equation}
where $\bm{y}_j$ include zeros except for $i_j$ entry, which includes $\bm{u}_{i_j}$. In other words, when the input is the correlation vector $\bm{D}^T\bm{r}$, the output is a matrix that contains $s$ columns, and each column $\bm{y}_j$ contains zeros everywhere except for index $i_j$ (which is the index of one of the $s$ most correlated atoms with the residual). The $\textit{MSPT}$ unit is a generalized version of $\textit{MPT}$, and in the case of $s=1$, the two are the same.

\subsubsection{S Atom Selecting (SAtoS) Unit}

Following the previous computational step, this unit extracts the $s$ selected atoms. Given the output of the $\textit{MSPT}$ function, $\bm{Y} \in \mathbb{R}^{m \times s}$, and the dictionary $\bm{D}$, this unit yields the $s$ atoms in $\bm{D}$ corresponding to the $s$ indices that contain the non-zero values in the columns of $\bm{Y}$. Equivalently, the $\textit{SAtoS}$ unit is defined using $\textit{AtoS}$ mentioned earlier as follows:
\begin{eqnarray}\label{eq:SAtoS}
\textit{SAtoS}\left ( \bm{D},\bm{Y} \right ) = \left [ \textit{AtoS}\left ( \bm{D},\bm{y}_1 \right ), \textit{AtoS}\left ( \bm{D},\bm{y}_2 \right ), \cdots , \textit{AtoS}\left ( \bm{D},\bm{y}_s \right )  \right ].
\end{eqnarray}
Note that the composition of the units $\textit{SAtoS}$ and $\textit{MSPT}$ provides the functionality of selecting the $s$ most correlated atoms with the residual $\bm{r}$, i.e., if $\left \{ \bm{d}_1,\bm{d}_2,\cdots,\bm{d}_s \right \}$ are indeed these atoms, then $\textit{SAtoS} \left (\bm{D},\textit{MSPT}\left ( \bm{D}^T \bm{r} \right ) \right ) = \left [ \bm{d}_1,\bm{d}_2,\cdots,\bm{d}_s \right ]$ or a permutation of them.

\subsubsection{Constructing the L-SP Architecture}
Similarly to the way the previous LGM architectures have been constructed, now we turn to describe a single iteration of SP as a computation graph. As mentioned earlier, the aggregated sub-dictionary $\bm{D}_{S_k}$ is passed between the different layers instead of the support $S_k$. L-SP single layer is described in Algorithm \ref{algo:LSP_LAYER}, starting by finding the $s$ most correlated atoms with the current residual, obtained by composing the $\textit{MSPT}$ and $\textit{SAtoS}$ units. These atoms are added to the temporal sub-dictionary $\bm{D}_{\widetilde{S}_k}$ alongside atoms from the previous layer. A representation under $\bm{D}_{\widetilde{S}_k}$ is calculated ($\widehat{\bm{\alpha}}_{\widetilde{S}_k}$). Next, $\bm{D}_{S_k}$ is calculated by finding the atoms with the $s$ maximum magnitudes in $\widehat{\bm{\alpha}}_{\widetilde{S}_k}$,  done by composing $\textit{MSPT}$ and $\textit{SAtoS}$ again. Using the updated support atoms $\bm{D}_{S_{k}}$, the representation $\widehat{\bm{\alpha}}_{S_{k}}$ is updated by solving the corresponding LS problem. Finally, the restored signal $\widehat{\bm{x}}_{k}=\bm{D}_{S_k}\widehat{\bm{\alpha}}_{S_k}$ is obtained.

Using the L-SP layer, the L-SP architecture is described in Algorithm \ref{algo:LSP_ARCH}. Like LGM, L-SP is also characterized by the dynamic number of layers that changes w.r.t. each input. The parameters of L-SP network are $\Theta = \left( \bm{D},s \right)$, where $\bm{D}$ is learned through back-propagation and $s$ is specified in advance, or somehow predicted in case it is not known. Similarly to the process described earlier, this scheme can be extended to include two dictionaries, regular and synthesis.

\begin{algorithm}[!htbp]
\SetAlgoLined
\SetKwFunction{FName}{LSPLayer}
\SetKwProg{Pn}{Function}{:}{}

  \Pn{\FName{$\bm{x} \in \mathbb{R}^{n}$, $\bm{D}\in \mathbb{R}^{n \times m}$, $\bm{D}_{S_{k-1}}\in \mathbb{R}^{n \times  s}$, $\widehat{\bm{x}}_{k-1} \in \mathbb{R}^{n}$}}{
  \BlankLine
  $\bm{r}_{k-1} = \bm{x}-\widehat{\bm{x}}_{k-1}$\\
  
  $\bm{u}=\bm{W}_{\bm{D}} \bm{D}^T\bm{r}_{k-1}$\\
 
        $\bm{D}_{tmp} =\textit{AStoS} \left (\bm{D},\textit{MSPT}\left ( \bm{u} \right ) \right )$    

$ \bm{D}_{\widetilde{S}_{k}} = \begin{bmatrix}
\bm{D}_{S_{k-1}}, & \bm{D}_{tmp}
\end{bmatrix}~~~$

% $\bm{D}_{S_{k}}^T \bm{D}_{S_{k}} \bm{\alpha}_{S_{k}} = \bm{D}_{S_{k}}^T \bm{x}$ 
$\widehat{\bm{\alpha}}_{\widetilde{S}_k}= \left(\bm{D}_{\widetilde{S}_{k}}^T\bm{D}_{\widetilde{S}_{k}}\right)^{-1}\bm{D}_{\widetilde{S}_{k}}^T\bm{x}$

$\bm{D}_{S_k} =\textit{AStoS} \left (\bm{D}_{\widetilde{S}_{k}},\textit{MSPT}\left ( \bm{W}_{\bm{D}_{\widetilde{S}_{k}}}^{-1} \widehat{\bm{\alpha}}_{\widetilde{S}_k}  \right ) \right )$  

$\widehat{\bm{\alpha}}_{S_k}= \left(\bm{D}_{S_k}^T\bm{D}_{S_k}\right)^{-1}\bm{D}_{S_k}^T\bm{x}$

$\widehat{\bm{x}}_{k} = \bm{D}_{S_{k}} \widehat{\bm{\alpha}}_{S_{k}}$
\BlankLine
  }
  {\KwRet  $\{\bm{D}_{S_{k}}$ , $\widehat{\bm{x}}_{k}\}$}
 
 \caption{L-SP Layer Inference}
 \label{algo:LSP_LAYER}
\end{algorithm}

\begin{algorithm}[!htbp]

\SetAlgoLined
\SetKwProg{Input}{Input:}{}{}
\SetKwProg{Output}{Output:}{}{}
\SetKwProg{Init}{Init}{}{}
\SetKwFor{For}{for}{do}{endfor}
\SetKw{Break}{break}
\SetKw{Endif}{endif}
\SetKwFunction{FName}{LSPLayer}

\Input{$\bm{x} \in \mathbb{R}^{n}$}{}
\Output{$\widehat{\bm{x}} \in \mathbb{R}^{n}$}{}

\Init{
 $\widehat{\bm{x}}_{0} = \bm{x}$, $\bm{D}_{S_{0}} = \textit{AStoS} \left (\bm{D},\textit{MSPT}\left (\bm{W}_{\bm{D}} \bm{D}^T\bm{x} \right ) \right )$}{}
\BlankLine
\For{$k=1,2,...$}{
$\{ \bm{D}_{S_{k}}~,~\widehat{\bm{x}}_{k}\}$  = \FName{$\bm{x}$, $\bm{D}$, $\bm{D}_{S_{k-1}}$, $\widehat{\bm{x}}_{k-1}$} \\
$\bm{r}_k = \bm{x}-\widehat{\bm{x}}_{k}$ \\
\uIf{$\left \| \bm{r}_k \right \|_2 > \left \| \bm{r}_{k-1} \right \|_2 $}{
$\widehat{\bm{x}} = \widehat{\bm{x}}_{k-1}$ \\
    \Break \\
  }{\Endif}
}

 \caption{L-SP Network Inference}
 \label{algo:LSP_ARCH}
\end{algorithm}

\subsection{LGM MMSE}
The Random OMP \cite{RandOMP} algorithm is a key component in turning a pursuit algorithm into a Minimum Mean Square Error (MMSE) estimator. This algorithm operates in a similar manner to OMP except for one critical difference, which is the way a new atom is chosen in each iteration. As mentioned earlier, OMP chooses the atom most correlated with the residual, whereas, Random OMP chooses the atom randomly, drawn from a distribution $A\cdot e^{c \left |\bm{W}_{\bm{D}}\bm{D}^T\bm{r} \right|}$, i.e. giving higher probability to larger projection values. The MMSE estimation is obtained by an averaging on the representation vectors from $T$ different Random OMP instantiations:
\begin{eqnarray}\label{eq:Rand_OMP_MMSE}
\widehat{\bm{\alpha}}_{MMSE}=\frac{1}{T}\sum_{i=1}^{T} \bm{\widehat{\alpha}}_i ,
\end{eqnarray}
in which $\left \{ \bm{\widehat{\alpha}}_1,\bm{\widehat{\alpha}}_2,...,\bm{\widehat{\alpha}}_T \right \}$ are the representation vectors that have been obtained by different Random OMP runs.

We turn to describe how to unfold the Random OMP algorithm. To do so, we need first to define the Random Maximal-Projection-Thresholding ($\textit{RMPT}$) unit. This unit is very similar to $\textit{MPT}$ defined earlier, except for one difference, which is the way the index of the only surviving entry $i_0$ is chosen. $\textit{RMPT}$ chooses $i_0$ index randomly with the normalized correlation vector (its input) the as its PDF, where entries with absolute values smaller than $\tau=0.8 \left \| \bm{u}  \right \|_\infty$ are nullified and are not taken into account. Consequently, the Random OMP method is unfolded just like OMP except for the use of $\textit{RMPT}$ unit instead of $\textit{MPT}$. LGM MMSE method is achieved by unfolding $T$ instantiations of Random OMP in parallel with shared parameters and a common input. In addition, we also add the result of the regular LGM network (i.e. MAP estimation). Finally the output of LGM MMSE is obtained by averaging these results.
% by averaging their results LGM MMSE output is obtained.

%nullifyin thr ?

\subsection{Batch-OMP Acceleration}
\label{subsec:Batch-OMP}
The Batch-OMP algorithm \cite{rubinstein2008efficient} is a method to accelerate the run-time of OMP when applied to a large batch of signals with the same dictionary $\bm{D}$. The main concept behind this method is to do as many computations as possible in advance, and since we are using the same dictionary for all the signals, these computations are shared between all the signals in the batch. More specifically, given a batch of signals to handle $\bm{X}=\left [ \bm{x}_1,\bm{x}_2,\cdots,\bm{x}_r \right ] \in \mathbb{R}^{n \times r}$, $\bm{D}^T\bm{D}$ and $\bm{D}^T\bm{X}$ are calculated in advance, and then the sparse representations are calculated for all of these signals using these pre-calculated matrices. Batch-OMP uses also a variation of Cholesky decomposition in order to solve the LS problem efficiently within the OMP. The authors of \cite{rubinstein2008efficient} have shown that if the batch is large, then Batch-OMP is more efficient in terms of run-time than the regular OMP. Thus, Batch-OMP can be used (with the corresponding adjustments) in order to accelerate the run-time inference of the LGM network in the evaluation part, where there is no need to calculate the derivatives. Moreover, some techniques used in Batch-OMP such as Cholesky decomposition can be also used in order to accelerate the inference of the LGM network in general (including training part). 
% instead of doing these calculations many times for different signals.
% (TODO Later), mention cholesky decomposition and the evaluation model.

\subsection{Learned-GCMP: CSC Based LGM}
\label{subsec:CSC}
\subsubsection{CSC Model $\And$ GCMP Pursuit Algorithm}

When dealing with high dimensional signals, applying the sparse prior becomes challenging. More specifically, in such cases the dictionary dimensions become gigantic, making it hard to store and almost impossible to multiply with in a pursuit algorithm. A popular method to tackle this disadvantage is to apply the sparse prior on local patches as discussed later in Section \ref{sec:LGM_var2}. Another method is to use the CSC model that presents a global signal model without suffering from the disadvantages of the global sparse prior.

The CSC model is a special case of the sparse model where $\bm{D}$ is a concatenation of $m$ banded circulant matrices, where each such matrix has a band of width $n \ll N$, in which $N$ is the dimension of the signal. As such, by a simple permutation of its columns, such a dictionary consists of all shifted versions of a local dictionary $\bm{D}_L$ of size $n \times m$ and contains $mN$ global atoms, i.e. $\bm{D} \in \mathbb{R}^{N \times mN}$ and the corresponding representation becomes a vector of the form $\bm{\alpha} \in \mathbb{R}^{mN}$. Under this structure, each patch of size $n$ in the signal is affected only by the atoms whose support overlap it. The subvector in $\bm{\alpha}$ that matches these atoms, is referred to as the \emph{stripe} of this patch. Since the dictionary $\bm{D}$ consists of all shifted versions of $\bm{D}_L$, each such stripe consists of $(2n-1)m$ entries. Note that overlapping patches are represented by overlapping stripes in the global sparse representation. For further information and analysis of the CSC model, the reader is referred to \cite{CSC1,CSC-GCMP,CSC2}.

GCMP \cite{CSC-GCMP} is a greedy pursuit algorithm that seeks to approximate the global representation vector under the assumption that the signal of interest can be modeled with the CSC model prior. This algorithm is inspired by the observation that when dealing with the CSC model, it is preferred to have a representation vector which is ``locally sparse'' rather than globally sparse \cite{CSC2}. As such, GCMP operates by initializing a zero representation vector, and then, at each iteration, the ``local sparsity'' in the above-mentioned stripes is increased by one.

\subsubsection{Unfolding GCMP}
Similarly to the process which have been done earlier, we start by presenting the core units of the construction, and then unfolding GCMP. Like the previous methods, $\bm{D}^T\bm{r}$ and $\bm{D}\widehat{\bm{\alpha}}$ are calculated several times throughout the deployment of GCMP or its learned version. However, since we are dealing with high dimensional signals, calculating these expressions directly using matrix multiplication is expensive in run-time, and storing $\bm{D}$ itself is also challenging. However, since $\bm{D}$ consists of all shifted versions of $\bm{D}_L$, we can store $\bm{D}_L$ instead. Moreover, calculating $\bm{D}^T\bm{r}$ and $\bm{D}\widehat{\bm{\alpha}}$ can be done efficiently using convolutions operation with the columns of $\bm{D}_L$ (or their flipped version). Analogously to the $\textit{MPT}$ unit, $\textit{GMPT}$ unit is defined in Algorithm \ref{algo:GMPT}, and is responsible for choosing which atoms are added at each iteration alongside their coefficients. For atom number $i_0$, $\Omega_{i_0}$ is the group of indices of atoms whose support overlap atom number $i_0$. Using the $\textit{GMPT}$ unit, GCMP algorithm is unfolded into the L-GCMP network, and its inference is described in Algorithm \ref{algo:LGCMP}.

\begin{algorithm}[!htbp]
\SetAlgoLined
\SetKwFunction{FName}{GMPT}
\SetKwProg{Pn}{Function}{:}{}
\SetKwFor{While}{while}{do}{endwhile}

  \Pn{\FName{$\bm{u} \in \mathbb{R}^{mN}$}}{
  \BlankLine
  $\bm{y} = \bm{0} \in \mathbb{R}^{mN}$
  
  \While{$\max_i \left | u_i \right | > 0$}{
  $i_0=\arg\max_i \left | u_i \right |$
  
  $\bm{y}_{i_0}=\bm{u}_{i_0}$
  
  $\bm{u}_{\Omega_{i_0}}=\bm{0}$
  
  }
  
  \BlankLine
  }
  {\KwRet  $\bm{y}$}
 
 \caption{Group-Maximal-Projection-Thresholding}
 \label{algo:GMPT}
\end{algorithm}

\begin{algorithm}[!htbp]

\SetAlgoLined
\SetKwProg{Input}{Input:}{}{}
\SetKwProg{Output}{Output:}{}{}
\SetKwProg{Init}{Init}{}{}
\SetKwFor{For}{for}{do}{endfor}
\SetKw{Break}{break}
\SetKw{Endif}{endif}
\SetKwFunction{FName}{LGMLayer}

\Input{$\bm{x} \in \mathbb{R}^{N}$}{}
\Output{$\widehat{\bm{x}} \in \mathbb{R}^{N}$}{}

\Init{
  $\bm{r}_{0} = \bm{x}$ , $\widehat{\bm{\alpha}}_{0} = \bm{0} \in \mathbb{R}^{mN}$}{}
\BlankLine
\For{$k=1,2,...,s$}{
$\widehat{\bm{\alpha}}_k =\widehat{\bm{\alpha}}_{k-1} +\bm{W}_{\bm{D}} \textit{GMPT}\left ( \bm{W}_{\bm{D}}\bm{D}^T\bm{r}_{k-1} \right) $ \\
$\widehat{\bm{x}}_{k} = \bm{D}\widehat{\bm{\alpha}}_k$ \\
$\bm{r}_k = \bm{x}-\widehat{\bm{x}}_{k}$\\
\uIf{$\left \| \bm{r}_k \right \|_2 \leq \epsilon$}{
    \Break \\
  }{\Endif}
}
$\widehat{\bm{x}} = \widehat{\bm{x}}_{k}$

 \caption{L-GCMP Network Inference}
 \label{algo:LGCMP}
\end{algorithm}

\section{Synthetical Experiments}
\label{sec:syn_experiements}
In this section we describe the conducted synthetical experiments, in which, we compare the denoising performance of the proposed LGM network and some of its variants with LISTA. These experiments demonstrate the superiority of the LGM based networks over LISTA, which is known to be the state-of-the-art sparse coding oriented deep neural network prior to this work. 

% Each experiment starts by generating the synthetical data as will explained later, next we contaminate these signals by an additive white Gaussian noise, and then we compare the denoising results between the proposed LGM net and LISTA.

$\textbf{Data Generation}$: First we build the dictionary $\bm{D}$ to be the DCT matrix of size $100 \times 400$, by sampling the cosine wave in different frequencies. Next we generate the representation vector $\bm{\alpha} \in \mathbb{R}^{400}$ for each signal, setting the cardinality to be $s$, and choosing the location of these $s$ non-zeros in $\bm{\alpha}$ randomly. The absolute value of each non-zero coefficient is distributed Uniformly in the interval $(0,1]$, and its sign is also chosen randomly. After, each signal is created by multiplying its corresponding representation with the dictionary $\bm{D}$ (i.e. $\bm{D\alpha}$), and finally the signals are normalized by their $L_\infty$ norm.

$\textbf{Compared Methods}$: For each noise level we compare the denoising performance of these different methods:
\begin{itemize}
  \item LGM: We use the LGM version in which the synthesis dictionary is free from the analysis one, and both dictionaries are of the same shape as the true dictionary. We set $\epsilon=\sigma_{noise}\sqrt{n}$ \footnote{This is the square root of the expectation of $L_2$ norm square of the noise vector} (stopping criteria coefficient) and $s=15$ (maximum number of layers/cardinality).
  \item LGM Post-training MMSE: Applying LGM MMSE network with the parameters learned by the regular LGM network as its parameters. We set $t=5$ (number of unfolded Rand-OMP instantiations) and $\epsilon,s$ get the same values as LGM.
  \item LGM MMSE: LGM MMSE network trained from scratch. We set $t=5$ (number of unfolded Rand-OMP instantiations) and $\epsilon,s$ get the same values as LGM.
  \item LGM True Cardinality: LGM version in which the stopping criterion is the true cardinality/sparsity of the input signal, i.e. the number of unfolded layers of LGM in each signal equals its true sparsity.
  \item LISTA: We use LISTA model with $T=7$ (number of unfolded layers). Instead of using the LISTA version explained earlier we use the version in which the learned parameters are: $\bm{W}=\frac{1}{c}\bm{D}^T$, $\bm{D}_1=\bm{D}$, $\bm{D}_2=\bm{D}$ (synthesis dictionary) and $\bm{\theta}$. The reason behind this decision is to compare LISTA's learned dictionaries with the true one.
  \item OMP True Dict $\And$ Cardinality: Applying OMP algorithm given the true dictionary and the true cardinality of each input signal.
  \item OMP True Dict: Applying OMP algorithm given the true dictionary, this method uses the residual energy threshold stooping criteria (like LGM). We set $\epsilon,s$ as the same values as LGM.
  \item OMP True Dict MMSE: Applying LGM MMSE given the true dictionary.
  \item Oracle: Recovering each signal given the true dictionary $\bm{D}$ and its true support $S$, i.e. the Oracle restored version of a noisy signal $\bm{x}$ is $\widehat{\bm{x}}=\bm{D}_{S}\left(\bm{D}_{S}^T\bm{D}_{S}\right)^{-1}\bm{D}_{S}^T\bm{x}$.
\end{itemize}

$\textbf{Experiments Process}$: We conduct two experiments, in the first one we generate $10000$ training signals and $2000$ test signals, both with cardinality $10$. In the second experiment we generate $10000$ signals for each cardinality in the group $\left \{ 5,6,7,8,9,10 \right \}$, then we combine them together forming the training set ($60000$ signals in total). The test set for experiment 2 is created in the same manner as in its training ($12000$ signals in total). For each experiment, we sweep over the following noise levels (standard deviation) $\left \{ 0.04,0.06,0.08,0.1,0.12,0.14 \right \}$, then for each one of them we create the input-output training pairs by contaminating each signal by an additive white Gaussian noise with the chosen standard deviation. 

We initialize the learned models (LGM, LGM MMSE, LGM True Cardinality, LISTA) from the same random dictionary, and trained using the input-output training pairs defined earlier, seeking to minimize the $L_2$ loss function. For the LGM based models, we add to the loss expression a regularization term of the mutual coherence of the learned dictionaries\footnote{The mutual coherence of a dictionary $\bm{D} \in \mathbb{R}^{n \times m}$ is defined as: $\mu(\bm{D})=\max_{i \neq j} \frac{\left | \bm{d}_i^T\bm{d}_j \right |}{\left \| \bm{d}_i \right \|_2 \left \| \bm{d}_j \right \|_2}$. Informally, the mutual coherence indicates the maximum amount of ``shared'' information between two different atoms. This figure also plays a crucial role in many of the pursuit algorithms stability guaranties, more specifically, the smaller this figure can be, the larger the possibility of OMP to recover the true support. For more information about the mutual coherence the reader is referred to \cite{MikiBook}.}. More specifically, the loss function of LGM based models is:
\begin{eqnarray}\label{eq:LGM_LOSS_Syn}
\mathcal{L}= \left(\sum _{\left ( \bm{x},\bm{x}^* \right ) \in \textit{training set}} {\left \| \bm{x}^* - \widehat{\bm{x}} \right \|_2^2} \right ) + \xi\left ( \sum_{\bm{D} \in \textit{learned dictionaries}} \mu\left ( \bm{D} \right ) \right ),
\end{eqnarray}
in which $\bm{x}$ is the noisy signal, $\bm{x}^*$ is the clean signal, $\widehat{\bm{x}}$ is its denoised version (the output of the model), and $\xi$ is set to be $5e^{-5}$. All these models are trained using ADAM optimizer \cite{Kingma2015AdamAM} with batch size equals 50, the learning rate for LGM and LGM True Cardinality is $0.002$, $0.01$ for LGM MMSE and $0.00001$ for LISTA. 

\begin{figure}[!htbp]
    \centering
    \begin{subfigure}[b]{0.49\textwidth}
        \includegraphics[width=\textwidth]{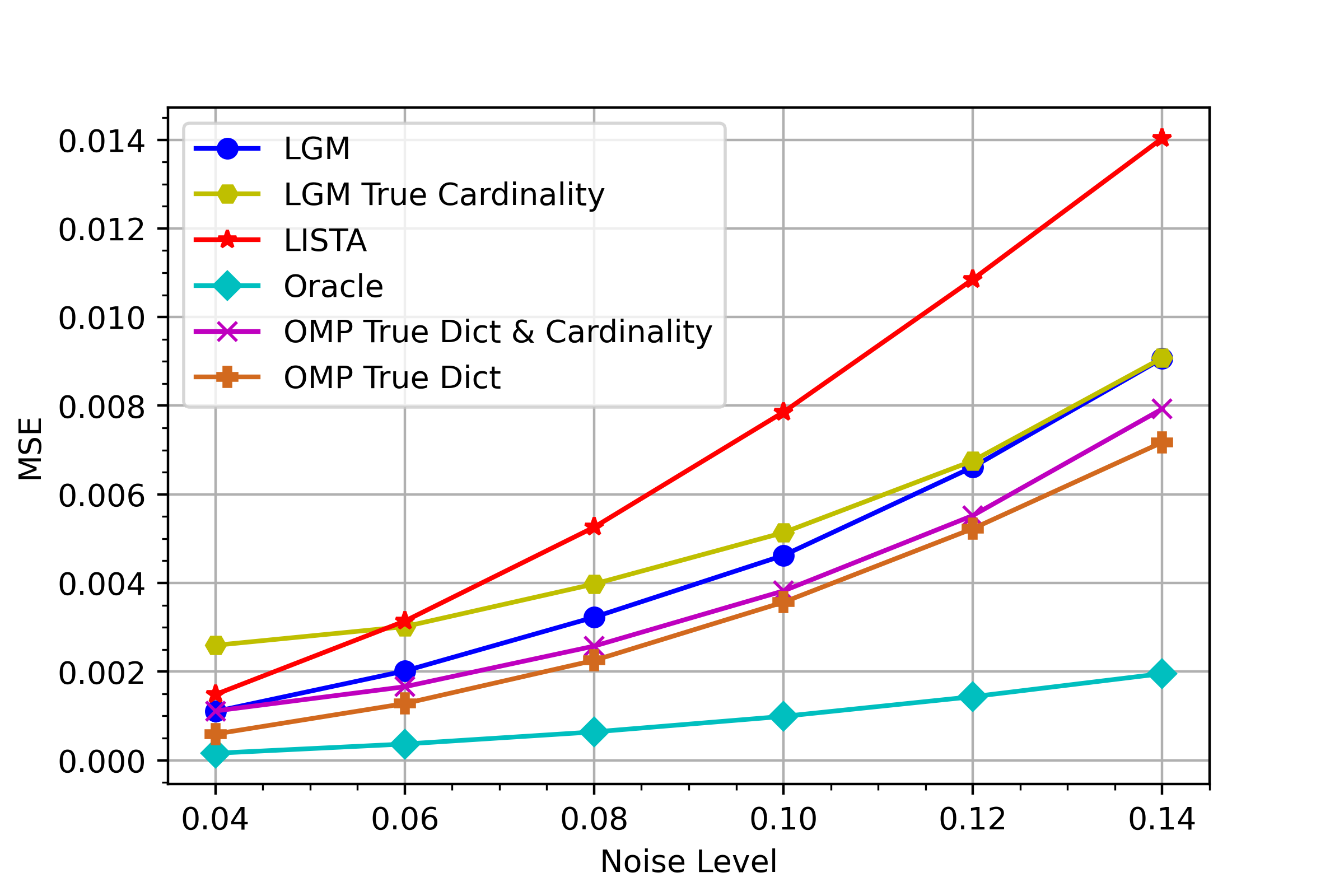}
        \caption{Group 1 MSE as a function of noise level.}
        \label{subfig:EX1-MSE1}
    \end{subfigure}
    \hfill
    \begin{subfigure}[b]{0.49\textwidth}
        \includegraphics[width=\textwidth]{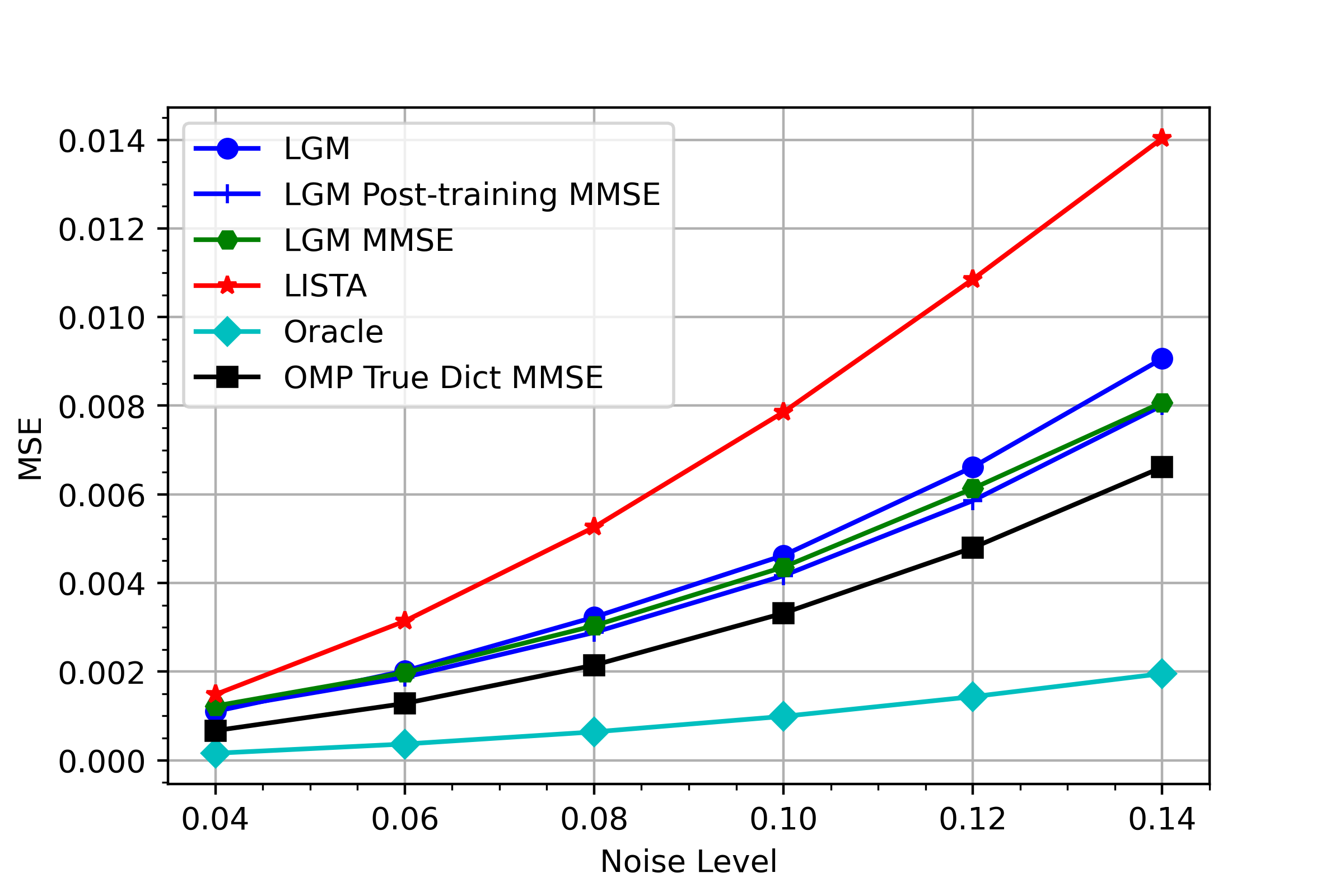}
        \caption{Group 2 MSE as a function of noise level.}
        \label{subfig:EX1-MSE2}
    \end{subfigure}
    
    \begin{subfigure}[b]{0.49\textwidth}
        \includegraphics[width=\textwidth]{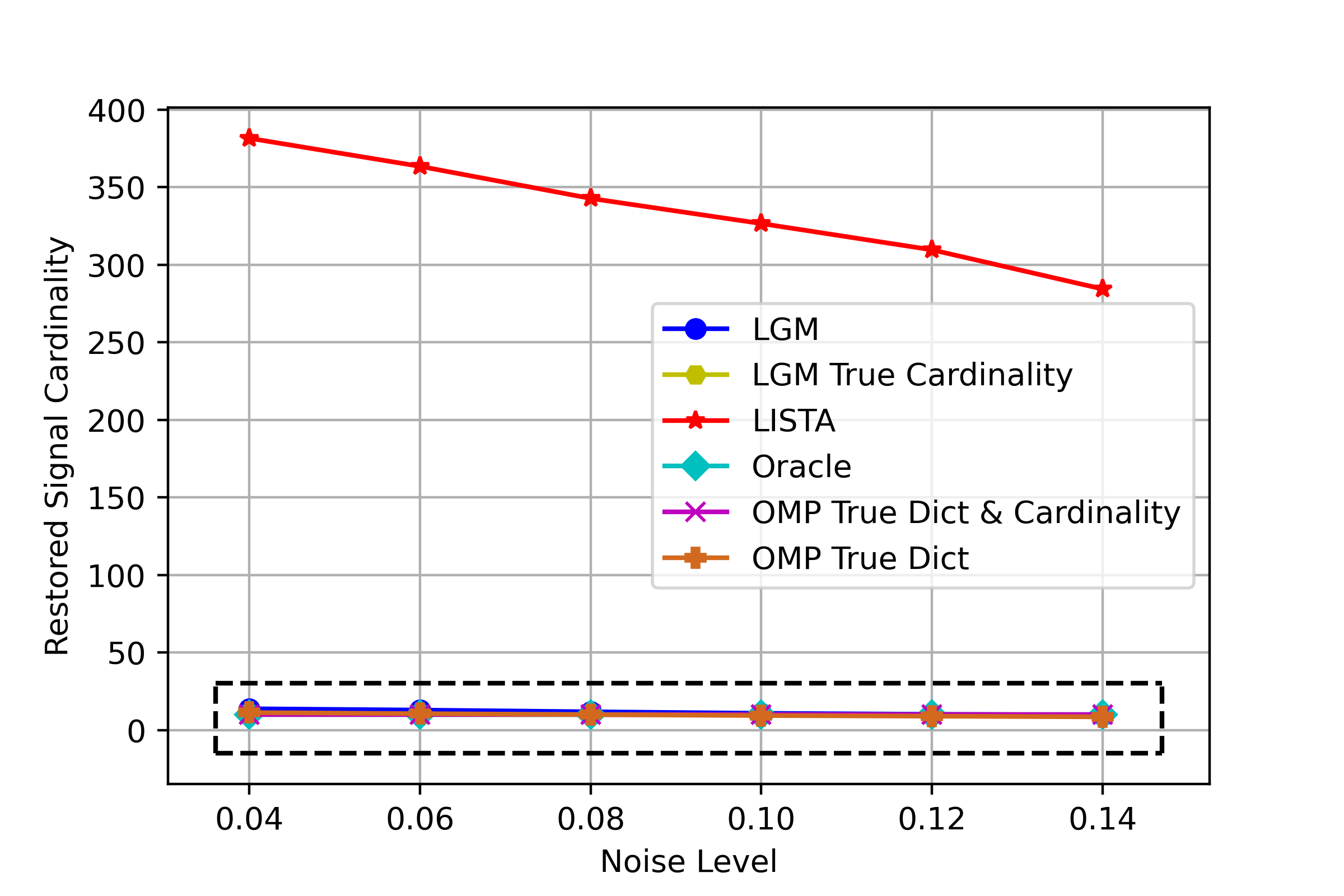}
        \caption{Group 1 average cardinality of the restored  signals as a function of noise level.}
        \label{subfig:EX1-sparsity1}
    \end{subfigure}
    \hfill
    \begin{subfigure}[b]{0.49\textwidth}
        \includegraphics[width=\textwidth]{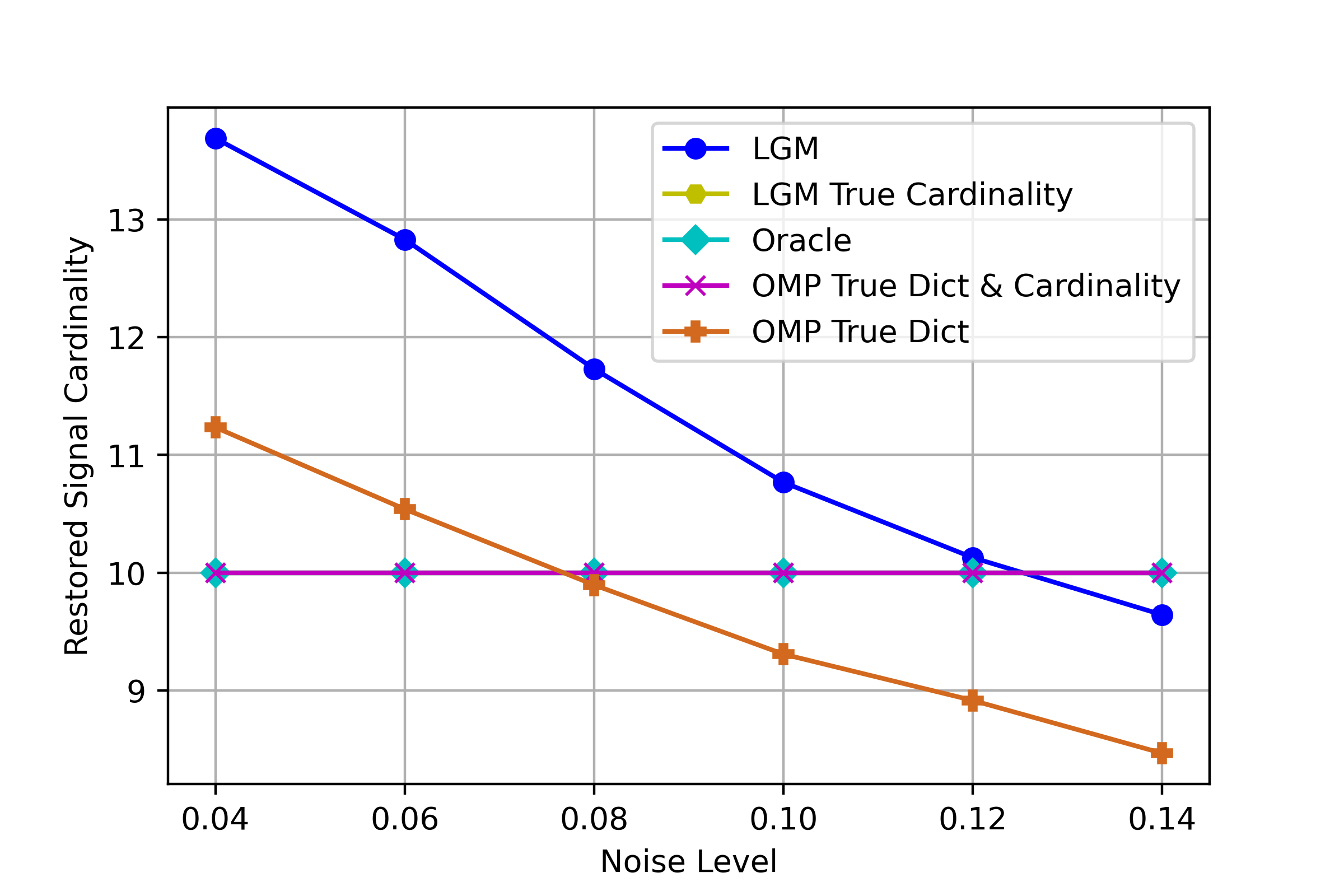}
        \caption{Group 1 average cardinality of the restored  signals zoom-in}
        \label{subfig:EX1-sparsity1-zoom}
    \end{subfigure}
    \caption{Synthetic experiment 1 (true cardinality $10$) testset results.}
    \label{fig:Synthitic_1_Res}
\end{figure}

\begin{figure}[!htbp]
    \centering
    \begin{subfigure}[b]{0.49\textwidth}
        \includegraphics[width=\textwidth]{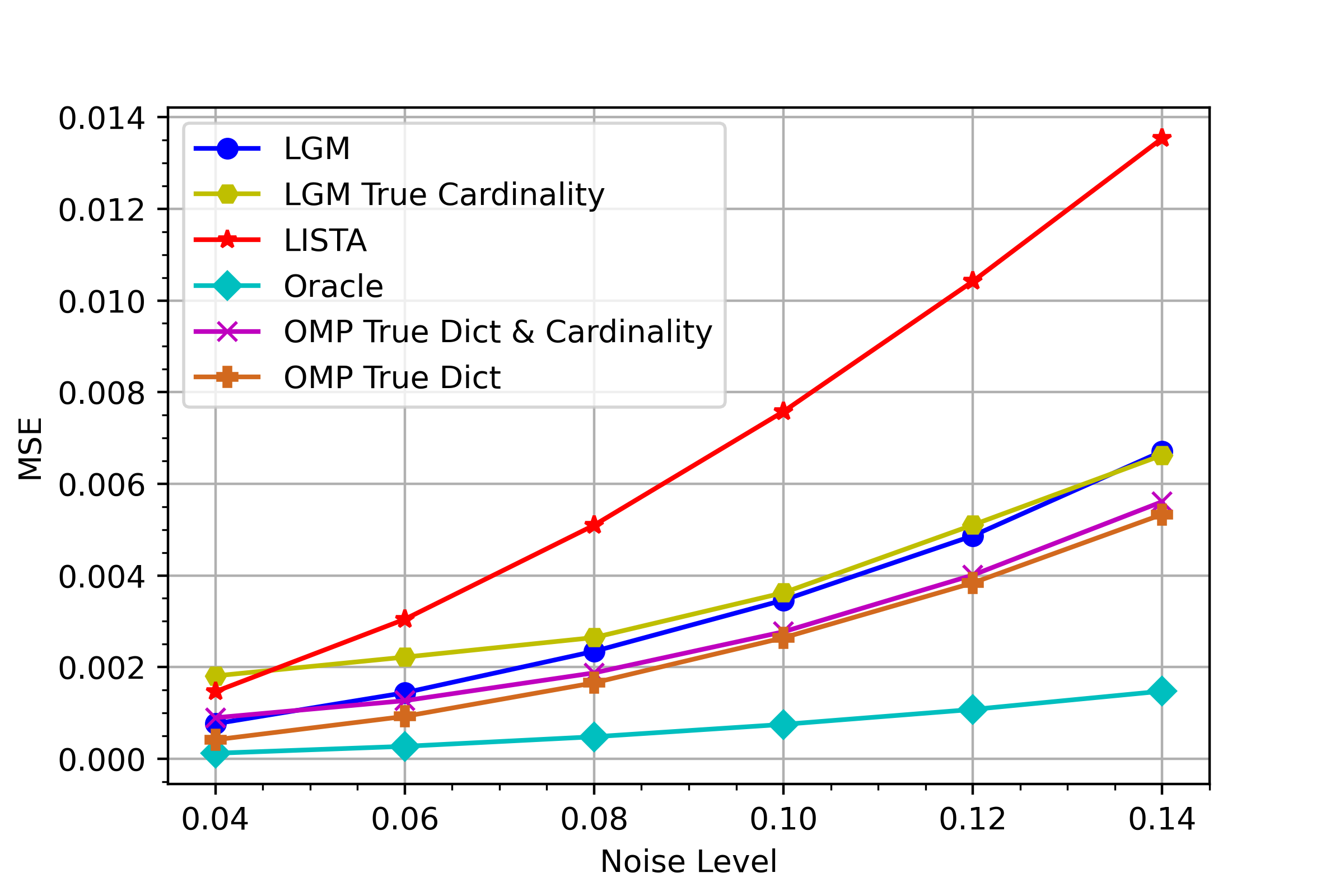}
        \caption{Group 1 MSE as a function of noise level}
        \label{subfig:EX2-MSE1}
    \end{subfigure}

    \begin{subfigure}[b]{0.49\textwidth}
        \includegraphics[width=\textwidth]{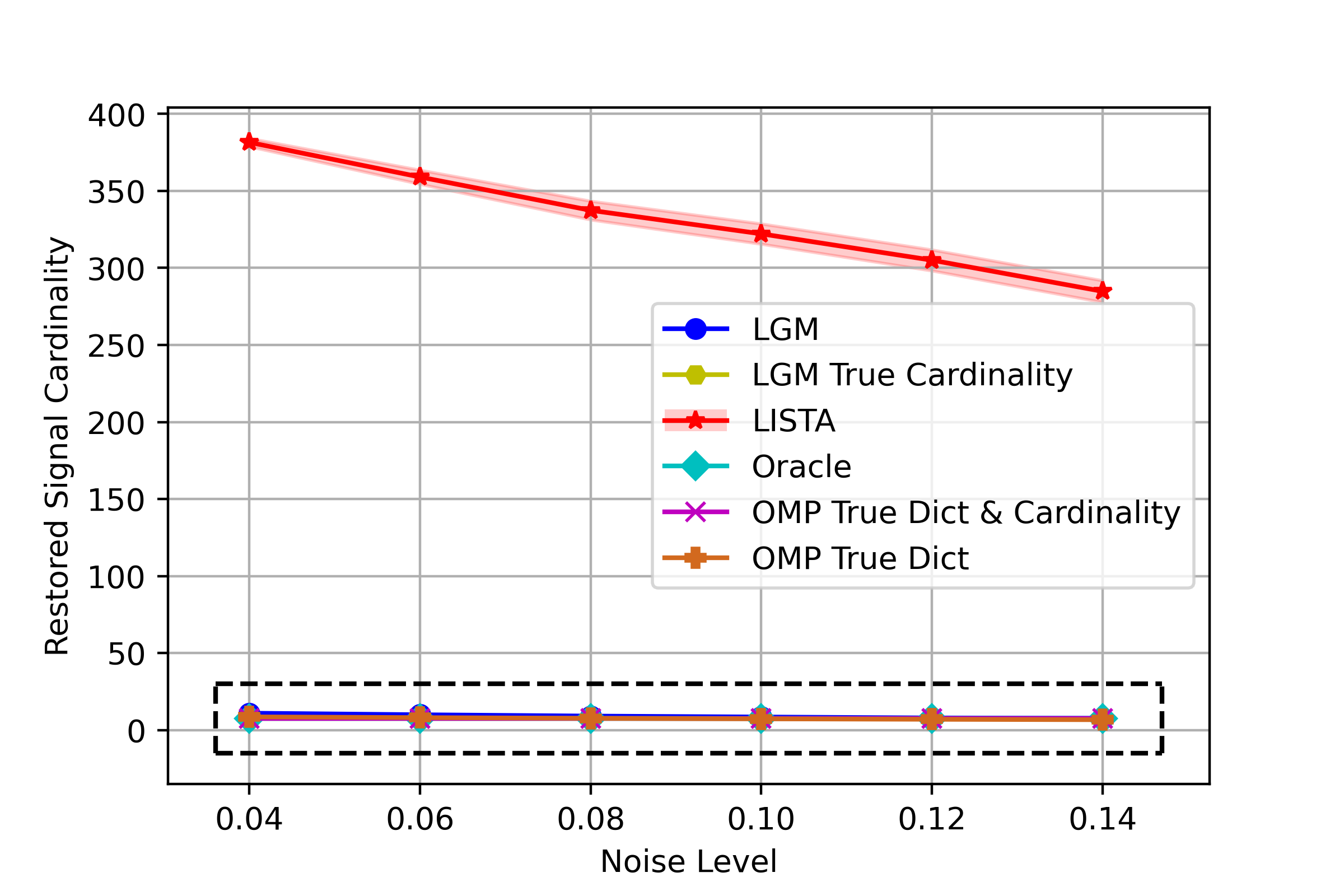}
        \caption{Group 1 average cardinality of the restored signals as a function of noise level. The area marked with light red around LISTA is the standard deviation of its restored cardinality.}
        \label{subfig:EX2-sparsity1}
    \end{subfigure}
    \hfill
    \begin{subfigure}[b]{0.49\textwidth}
        \includegraphics[width=\textwidth]{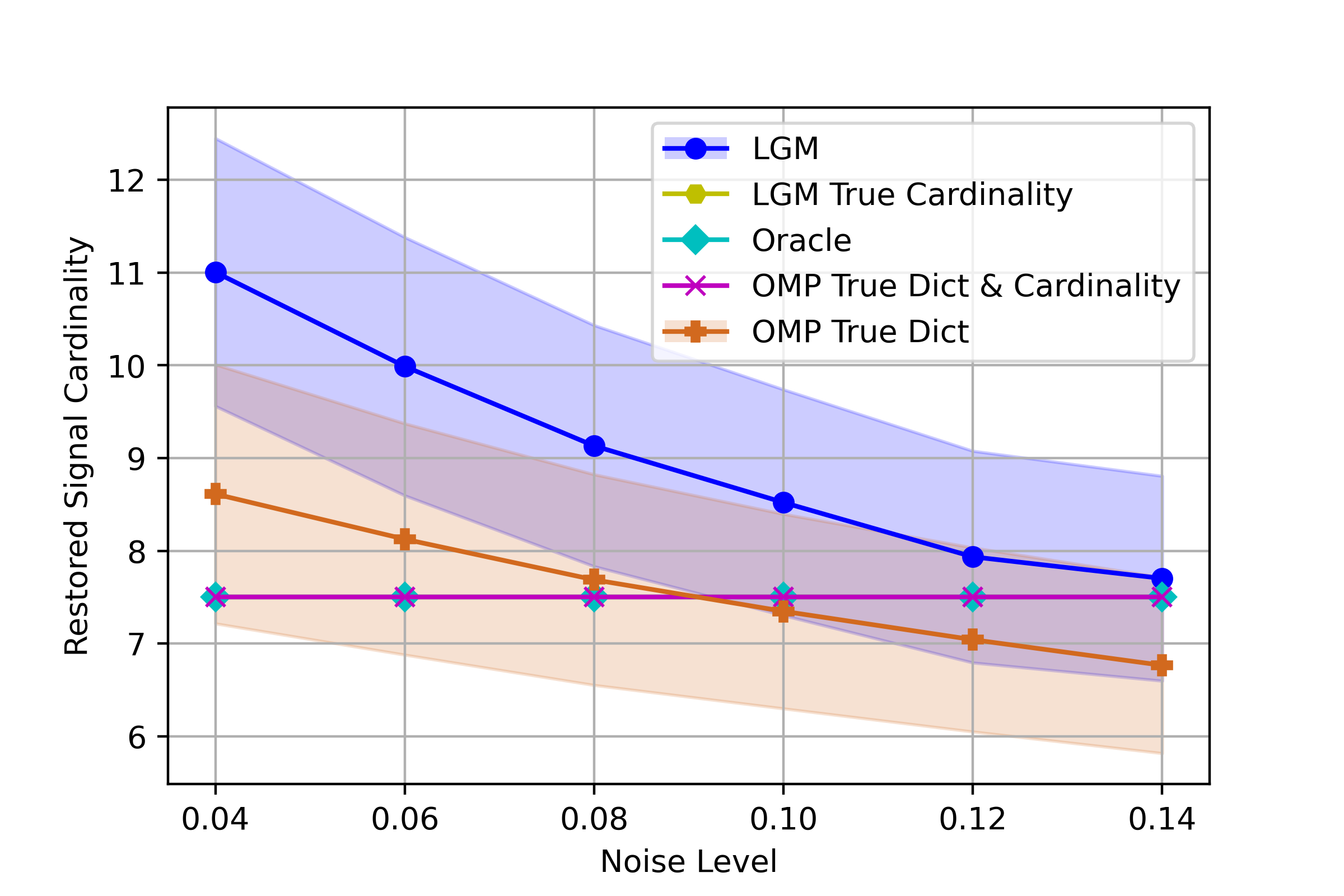}
        \caption{Group 1 average cardinality of the restored signals zoom-in. The areas marked with light blue and chocolate around LGM and OMP True Dict respectively are the standard deviation of there restored cardinality.}
        \label{subfig:EX2-sparsity1-zoom}
    \end{subfigure}
    \caption{Synthetic experiment 2 (true cardinality $5-10$) testset results.}
    \label{fig:Synthitic_2_Res}
\end{figure}

$\textbf{Results}$: In order to have a clearer comparison, we split the compared methods into two (overlapping) comparison groups, the first one contains LGM, LGM True Cardinality, LISTA, OMP True Dict $\And$ Cardinality, OMP True Dict and Oracle, the second one contains LGM, LGM Post-training MMSE, LGM MMSE, LISTA, Oracle and OMP True Dict MMSE. Figure \ref{fig:Synthitic_1_Res} presents the results of experiment 1 (true cardinality $10$), and Figure \ref{fig:Synthitic_2_Res} presents the results of experiment 2 (true cardinality $5-10$). Referring to the first group, Figures \ref{subfig:EX1-MSE1} and \ref{subfig:EX2-MSE1} describe the MSE performance on the testset as a function the noise level, and it is clearly observed that LGM outperforms LISTA and has a relatively close performance to OMP True Dict. Figures \ref{subfig:EX1-sparsity1}, \ref{subfig:EX1-sparsity1-zoom} and \ref{subfig:EX2-sparsity1}, \ref{subfig:EX2-sparsity1-zoom} describe the average cardinality of the restored testset signals, and as can be observed, LGM restored signals are much sparser than LISTA's ones (which are dense) and their recovered cardinality is very close to the true one. Referring to the second group, Figure \ref{subfig:EX1-MSE2} describes the MSE performance on the testset as a function the noise level, and the main observation from it is that, as expected, the MMSE approach gives a boost to the denoising performance. We did not test the MMSE approach in the second experiment since it is not our main focus.

\begin{figure}[!htbp]
    \centering
    \begin{subfigure}[b]{0.49\textwidth}
        \includegraphics[width=\textwidth]{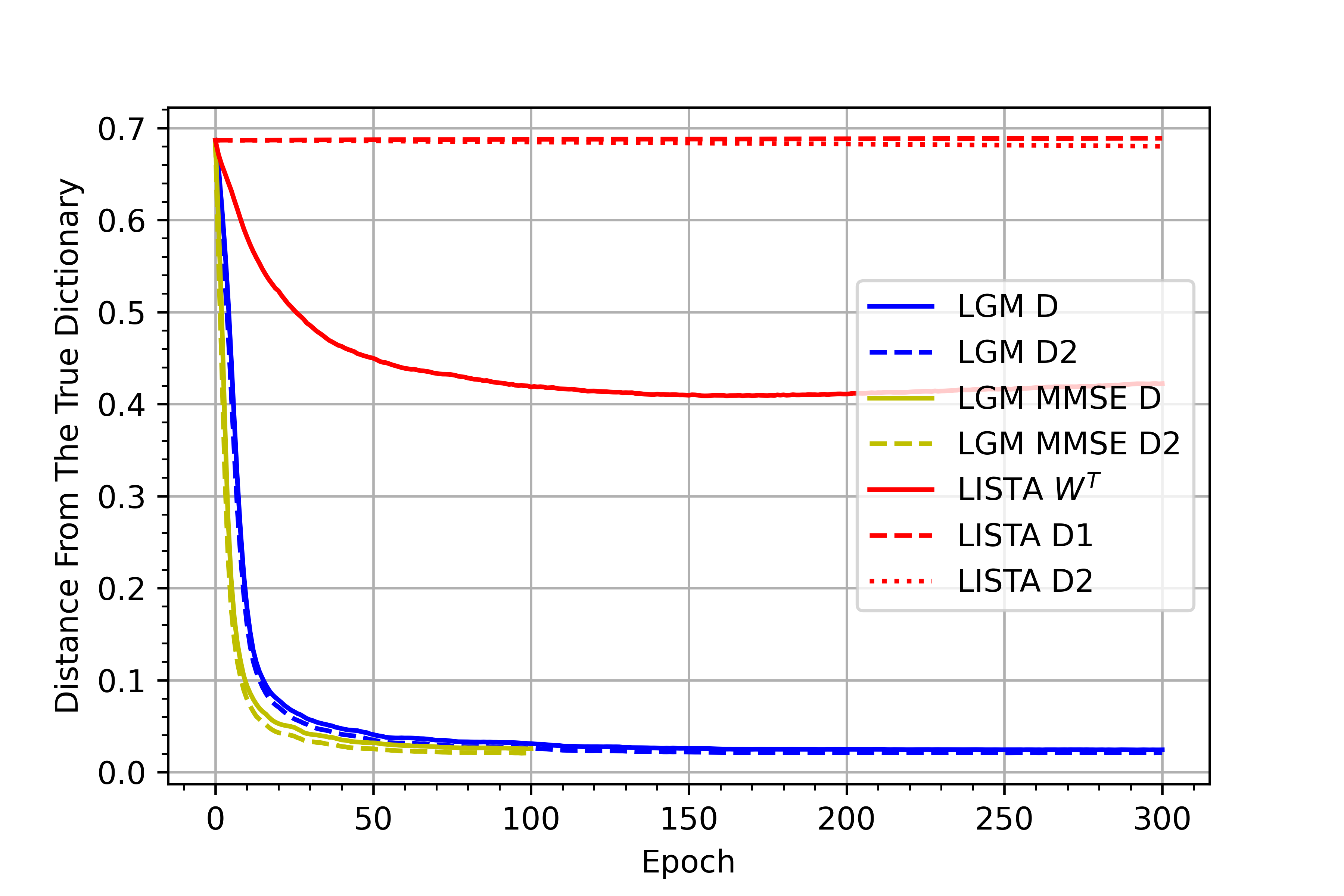}
        \caption{Distance from true dictionary.}
        \label{subfig:D_Dist_Per_Epoch}
    \end{subfigure}
    \hfill
    \begin{subfigure}[b]{0.49\textwidth}
        \includegraphics[width=\textwidth]{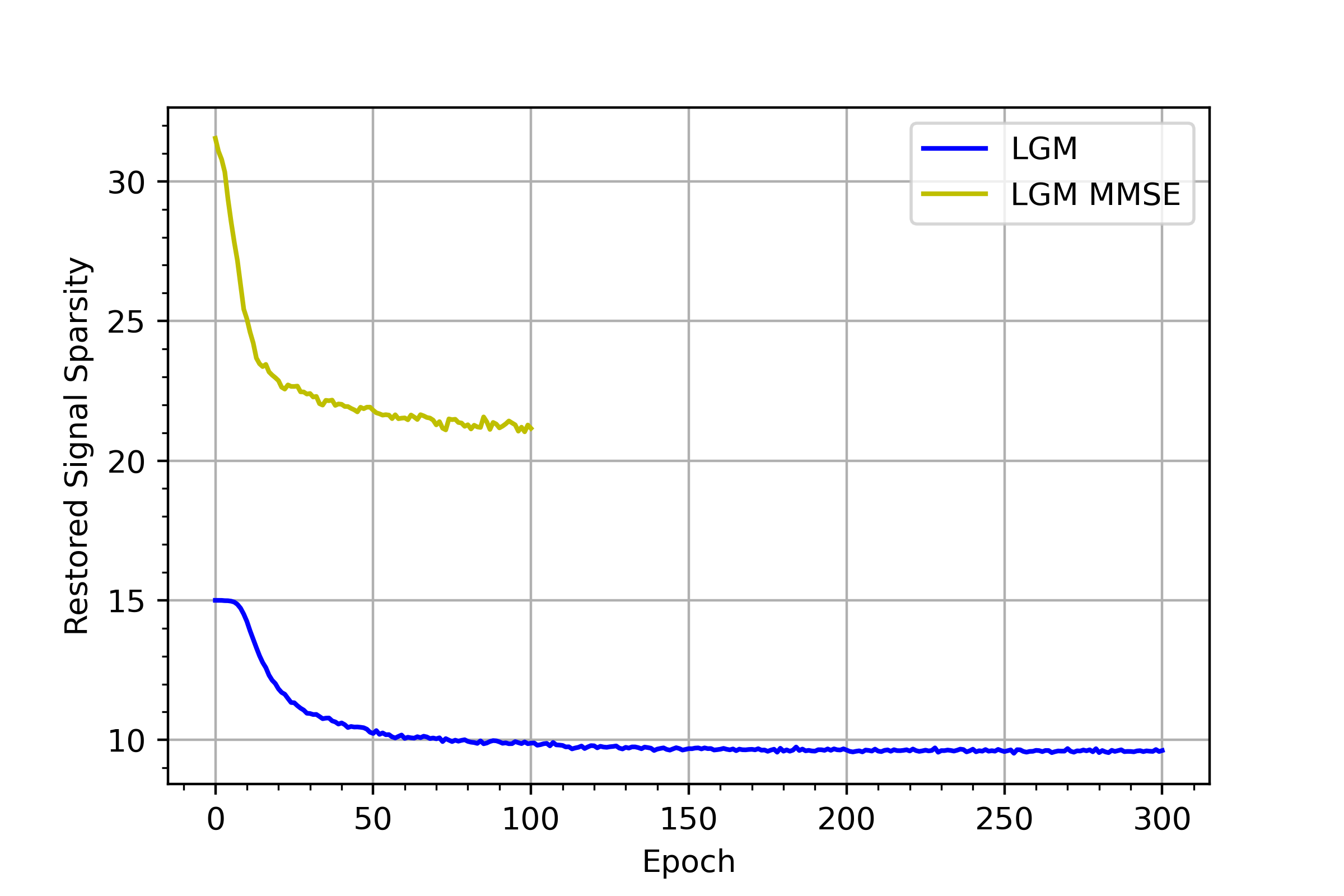}
        \caption{Average cardinality of the testset restored signals.}
        \label{subfig:Sparsity_Per_Epoch}
    \end{subfigure}
    \caption{Distance from the true dictionary and average cardinality during training on a specific noise level in synthetic experiment 1 (true cardinality $10$)}
    \label{fig:Synthitic_1_Extra}
\end{figure}

\begin{figure}[!htbp]
    \centering
    \includegraphics[width=\textwidth]{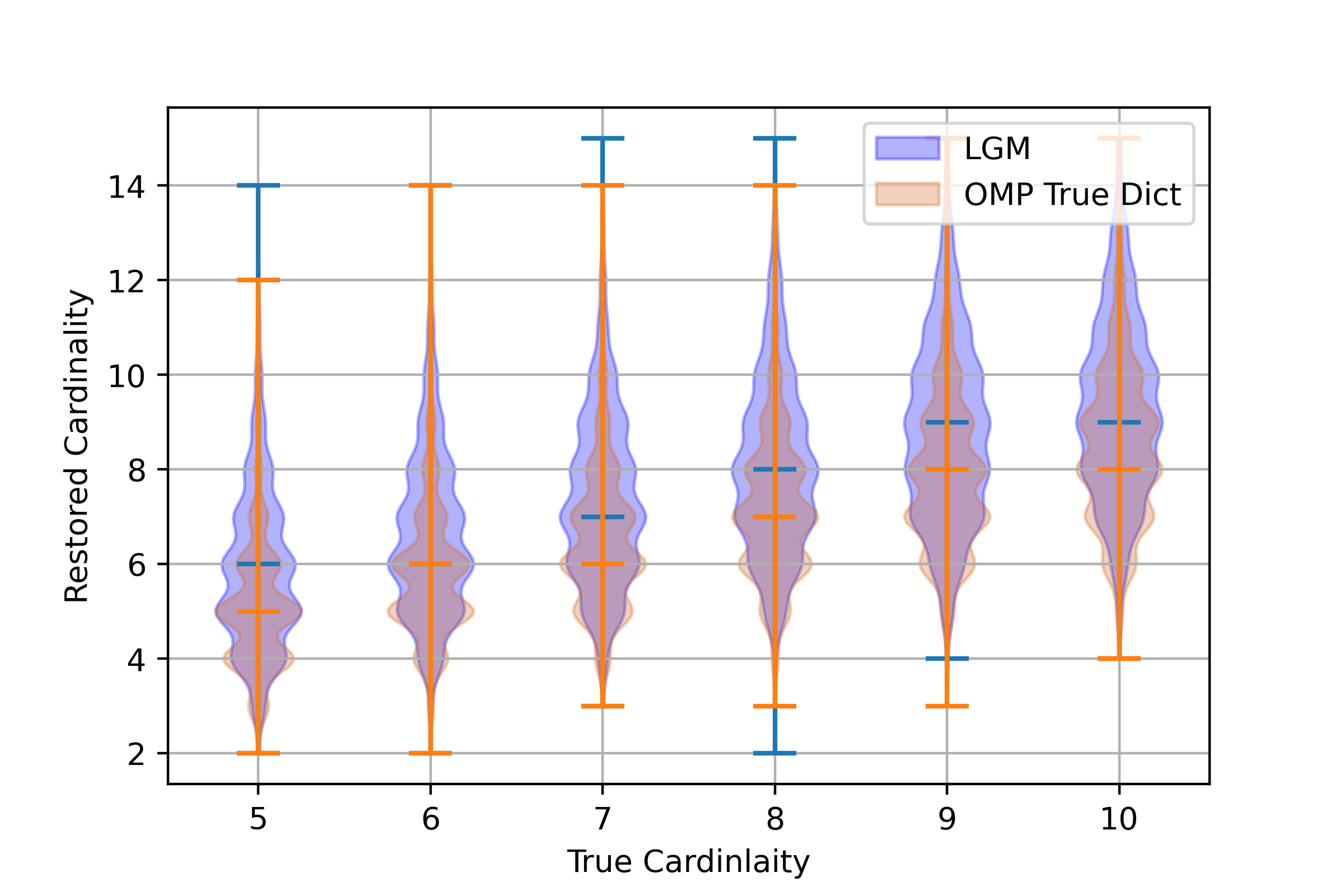}
    
    \caption{The distribution of the restored cardinality as a function of the true cardinality on a specific noise level in experiment 2 (true cardinality $5-10$)}
    \label{fig:Synthitic_2_Extra}
\end{figure}

Figure \ref{fig:Synthitic_1_Extra} presents the learned dictionaries distance from the true one, and the average cardinality of the restored signals in the test set during training, a specific noise level. The dictionary distance metric that we use is defined in the Supplementary Materials \ref{sub_section:Dict_Dist}, this metric values lie in the interval $[0,1]$, and the smaller it gets, the closer the dictionaries are. As observed in this figure, LGM and LGM MMSE learned dictionaries almost converge to the true one, while LISTA's learned ones are far away from it. Moreover, Figure \ref{fig:Synthitic_2_Extra} presents the distribution of the cardinality of the restored test set signals on a specific noise level in experiment 2. The results of these experiments clearly demonstrate the superiority of the LGM method compared to LISTA when dealing with true sparse data. These experiments alongside the following experiments are implemented in \url{https://github.com/RajaeeKh/LearnedGreedyMethod-LGM}.

% Figure \ref{fig:Synthitic_MSE} describes the denoising Mean Squared Error (MSE) performance on the test set as a function of the noise level for both models, and as can be noticed, LGM outperforms LISTA in almost all of the noise levels.

% Then for each noise level (standard deviation), both models are initialized all over again and then trained as denoisers on the training samples minimizing the error $l_2$ loss function between the denoised signal and the clean one. Both models are trained using ADAM optimizer \cite{Kingma2015AdamAM} with batch size equals 50 and $0.001$ learning rate for LGM and $0.0001$ for LISTA. Figure \ref{fig:Synthitic_MSE} describes the denoising Mean Squared Error (MSE) performance, and as can be noticed, LGM outperforms LISTA in almost all of the noise levels.

\section{LGM in Image Processing Applications}
\label{sec:LGM_var2}
\subsection{LGM for Image Denoising}
We move now to the image denoising problem, in which our task is to recover a clean image $\bm{x}^* \in \mathbb{R}^N$ given its noisy version $\bm{x} \in \mathbb{R}^N$, i.e. $\bm{x} = \bm{x}^*+\bm{v}$, where $\bm{v} \in \mathbb{R}^N$ is an additive white Gaussian noise vector with zero mean and standard deviation $\sigma$. Following \cite{K-SVD}, when dealing with natural images, the sparse prior model can be imposed on the image patches instead of the whole image. More specifically, for the image denoising task, the noisy image $\bm{x}$ is divided into fully overlapping $p \times p$ patches, and then each patch undergoes a pursuit operation in order to obtain an approximate representation $\widehat{\bm{\alpha}}$ under a given dictionary $\bm{D}$. Next, the restored patches are synthesized using their representation, and finally the restored image is created by returning each restored patch to its original location and averaging over the overlaps. More formally, returning the restored patches into their original location requires solving the following problem:
\begin{eqnarray}\label{eq:sparse_denoising1}
\widehat{\bm{x}}=\arg\min_{\bm{y}} ~ \lambda \left \| \bm{x} - \bm{y} \right \|_2^2+\sum_i\left \| \bm{R}_i \bm{y} - \bm{D}\widehat{\bm{\alpha}}_i \right \|_2^2,
\end{eqnarray}
where $\bm{R}_i \in \mathbb{R}^{p^2 \times N}$ is a matrix that extracts the $i-th$ patch from the image, and $\lambda$ is a scalar parameter that is related to the noise level. This problem is a quadratic problem and its closed form solution is
\begin{eqnarray}\label{eq:sparse_denoising2}
\widehat{\bm{x}}=\left ( \lambda \bm{I} + \sum_i \bm{R}_i^T\bm{R}_i \right )^{-1}\left ( \lambda \bm{x}+\sum_i \bm{R}_i^T\bm{D}\widehat{\bm{\alpha}}_i \right ).
\end{eqnarray}

\noindent Now we present an LGM based denoising end-to-end architecture along the above lines.

\subsubsection{Basic LGM Denoising Architecture}
We start by describing the patch denoising part, and then we move on to describe the whole end-to-end image denoising architecture.

\noindent  $\textbf{Patch Denoising}$: This part is an LGM network that takes a noisy patch of size $p \times p$ and returns its restored version. As mentioned earlier, we use an LGM network in which the synthesis dictionary is freed from the analysis one, both dictionaries are initialized with DCT dictionary of size $p^2 \times 4p^2$. Following \cite{SparseMultiScale2007},  an atom of ones (up to a scalar) is added to each dictionary for better handling of the DC in each patch, and when calculating the correlation vector $\bm{u}$, the entry corresponding to it is not divided by its norm, i.e. $\bm{W}_{\bm{D}} = diag^{-1} \left ( \left \| \bm{d}_1  \right \|_2, \left \| \bm{d}_2  \right \|_2, ... , \left \| \bm{d}_{4p^2}  \right \|_2 ,1 \right )$. Each dictionary gets a different scalar, and both are initialized equally with value $2.5$, then they are learned during the learning process.

Instead of using a hard-coded residual threshold stopping criteria as in the original LGM architecture, we use a small fully-connected DNN to perform this task. Recall that we use LGM net with maximum cardinality $s$ (i.e. LGM net includes $s$ layers), and the output of layer number $i$ is $\widehat{\bm{x}}_i$ which is the restored signal until that layer. Thus, given an input signal $\bm{x}$, we apply the LGM scheme mentioned earlier, getting the output of all layers $\left \{ \widehat{\bm{x}}_1,\widehat{\bm{x}}_2,...,\widehat{\bm{x}}_s \right \}$ along with their corresponding residuals $\bm{r}_i=\bm{x}-\widehat{\bm{x}}_i$. These residuals are organised as the columns of a matrix of size $p^2 \times s$, fed to the small FC-DNN mentioned earlier, and its output $\bm{p} \in \mathbb{R}^s$ is a weights vector that includes the weight of each $\widehat{\bm{x}}_i$, such that the output signal is $\widehat{\bm{x}}=\sum_{i=1}^s p_i \widehat{\bm{x}}_i$. In other words, this small FC-DNN somehow models the error threshold stopping criteria by giving a higher weight to to the best residual, thus giving an attention to it. We call this an ``Attention Network'', and it is composed of 5 layers, the first 4 are fully connected layers defined by multiplying the input matrix with $\bm{W}_1 \in \mathbb{R}^{p^2 \times p^2}$ from the right and $\bm{W}_2 \in \mathbb{R}^{s \times s}$ from the left, these layers are followed by adding a bias then applying ReLU activation function. The last layer is defined by multiplying the input matrix with $\bm{W} \in \mathbb{R}^{p^2 \times 1}$ from the right and applying a Softmax function.

\noindent $\textbf{End-to-End Denoising}$: The noisy image $\bm{x}$ is broken into fully overlapping $p \times p$ patches. Each patch $\bm{x}_i$ undergoes the patch denoising scheme mentioned above, and the denoised image is reconstructed by the following equation:
% \begin{eqnarray}\label{eq:LGM_denoising}
% \widehat{\bm{x}}=\left ( \lambda \bm{I} + \sum_i \bm{R}_i^T\bm{R}_i \right )^{-1}\left ( \lambda \bm{x}+\sum_i \bm{R}_i^T\widehat{\bm{x}}_i \right ),
% \end{eqnarray}

\begin{eqnarray}\label{eq:LGM_denoising}
\widehat{\bm{x}}=\left (  \sum_i \bm{R}_i^T\bm{R}_i \right )^{-1}\left ( \sum_i \bm{R}_i^T\widehat{\bm{x}}_i \right ),
\end{eqnarray}
in which $\widehat{\bm{x}}_i$ is the denoised version of patch $\bm{x}_i$. 
% , and $\mu$ is a scalar parameter which is also learned via back-propagation. 
The number of learned parameters of the basic LGM denoising architecture are:
% \begin{eqnarray}\label{eq:Basic_LGM_Num_Parameters}
% \underbrace{2}_{\textit{two dictionaries}} \times \left (  \underbrace{4p^4}_{\textit{dictionary parameters}} + \underbrace{1}_{\textit{DC atom coefficient}}\right ) + \underbrace{1}_{\lambda} = 8p^4+3.
% \end{eqnarray}

\begin{eqnarray}\label{eq:Basic_LGM_Num_Parameters}
& \underbrace{2}_{\textit{two dictionaries}} \times \left (  \underbrace{4p^4}_{\textit{dictionary parameters}} + \underbrace{1}_{\textit{DC atom coefficient}}\right ) + \underbrace{4\left(p^4+s^2+s\right)+p^2}_{\textit{attention net}} = \\ \nonumber
\\ \nonumber
& 12p^4+4\left(s^2+s\right)+p^2+2 \approx 12p^4+4s^2. \nonumber
\end{eqnarray}

\subsubsection{Advanced LGM Denoising Architecture}
The advanced LGM denoising network is compound of $2$ basic LGM denoising networks with $p=8,12$ and $s=10,20$ respectively. This network operates in two phases, the first phase consists of feeding the noisy images to each one of the basic LGM denoisers independently, and in the second phase the $2$ denoised images are combined together by using a smart averaging deep neural network. More specifically, in order to get the value of pixel $ \left ( i,j  \right )$ at the final image, patches with size $5 \times 5$ at center $ \left ( i,j  \right )$ are extracted from the $2$ denoised images and organised in a $2 \times 5 \times 5$ tensor, then it is flattened into 1d vector and fed to a small fully-connected deep neural network, the output of which is a scalar value of the output pixel. In addition, the denoised images (i.e. output of the first phase) are reflection padded in order to get final denoised image with the same size. The averaging network includes $3$ fully connected layers with biases, each followed by a ReLU activation function. The size of these layers is $50 \times 50$, and the size of the final layer is $50 \times 1$ (without bias). The network also has two weighted skip connections, one from input to middle (size $50 \times 50$), and one from input to output (size $50 \times 1$). The number of learned parameters of this network is:
\begin{eqnarray}\label{eq:Advanced_LGM_Num_Parameters}
\underbrace{\sum _{p \in \left \{ 8,12 \right \}} {\left ( 12p^4+4s^2 \right )}}_{\textit{basic LGM denoisers}} + \underbrace{4\left (50^2\right )+5\left ( 50 \right )}_{\textit{averaging DNN}} = 310,234.
\end{eqnarray}

\subsection{LGM For Image Deraining}
Image deraining is the process of removing rain streaks from an image. The widely used rain model \cite{Derain1,Derain2,Derain3} assumes that the captured rainy image $\bm{x} \in \mathbb{R}^N$ is expressed as $\bm{x}=\bm{x}^*+\widetilde{\bm{s}}$, in which $\bm{x}^* \in \mathbb{R}^N$ is the clean image and $\widetilde{\bm{s}} \in \mathbb{R}^N$ is the rain streaks component. The main observation in which the LGM deraining scheme built-on is that the rain streaks component which we aim to remove is a structured noise, thus it can represented using the sparse prior as well. Similarly to the LGM image denoising architecture mentioned earlier, this scheme also operates locally on all overlapping $p \times p$ sized patches. Since this scheme is applied on RGB images, we initialize both dictionaries with the same random dictionary of size $3p^2 \times 9p^2$. As before, a ones atom is added to the dictionary, and since we are dealing with RGB images, each channel is multiplied by a different scalar, and each scalar is initialized to be $2.5$.

As mentioned earlier, our goal is to express the rain streaks component using the sparse prior, thus we aim to split the dictionary atoms into two groups, one responsible for the image content and the other responsible for the rain streaks component. We operate in a fully supervised mode of work in which we have clean images and rain steaks images. Our algorithm is posed as a network that operates on a combination of an image and rain, and the output is matched to the clean image and the rain streaks image. We leverage almost the same architecture as used in the denoising, with one main difference - a separator of the atoms to the two contents. In order to achieve this separation, a vector $\bm{\theta} \in \mathbb{R}^{9p^2}$ is added to the learned parameters, and $\textit{f}\left ( \bm{\theta} \right ) \in \left [ 0,1 \right ]^{9p^2}$ is a coefficients vectors that describes the image content percentage of each atom. $\textit{f}$ is an element-wise function defined as $\textit{f}\left ( \theta \right )=min\left (  max\left ( 0,\theta \right ),1 \right )$, and all elements of $\bm{\theta}$ are initialized randomly in the interval $\left [0,1\right]$. Consequently, the proposed LGM deraining scheme operates by unfolding the LGM network for each patch independently until the residual energy is almost zero. Then, given the representation at the final layer $\widehat{\bm{\alpha}}_{S_k}$ for patch number $i$ ($k$ is the last layer), the content part is obtained by $\widehat{\bm{x}}_i^c = \bm{D}_{S_k} \left ( \widehat{\bm{\alpha}}_{S_k} \cdot \textit{f} \left ( \bm{\theta}_{S_k} \right ) \right )$ and the rain streaks part is obtained by $\widehat{\bm{x}}_i^s = \bm{D}_{S_k} \left ( \widehat{\bm{\alpha}}_{S_k} \cdot \left ( \bm{1} - \textit{f} \left ( \bm{\theta}_{S_k} \right ) \right ) \right )$, in which $\cdot$ is element wise multiplication and $\bm{\theta}_{S_k}$ is a sub vector of $\bm{\theta}$ at the corresponding indices. Next, the image content part $\widehat{\bm{x}}^c$ is created by averaging the content part of each patch like (\ref{eq:LGM_denoising}), and the same applies for the rain streaks part $\widehat{\bm{x}}^s$. Hopefully, during the training procedure, most of the element in $\textit{f}\left ( \bm{\theta} \right )$ converge to either $0$ or $1$, thus getting the desired separation between content and rain atoms.

The above process leads to two versions of the derained image $\widehat{\bm{x}}^c$ and $\bm{x}- \widehat{\bm{x}}^s$. We obtain the final derained image by combining these two using a smart averaging deep neural network like before. More specifically, in order to get the value of pixel $ \left ( i,j  \right )$ at the final image, patches with size $5 \times 5$ at center $ \left ( i,j  \right )$ are extracted from these three images and organised in a $9 \times 5 \times 5$ tensor, this tensor is flattened into 1d vector and fed to a small fully-connected deep neural network, the output of which is a 3 dimensional vector which represents the RGB components of the output pixel. In addition, the images ($\widehat{\bm{x}}^c$, $\bm{x}- \widehat{\bm{x}}^s$ and $\bm{x}$) are padded in order to get final denoised image with the same size. The averaging network includes $4$ fully connected layers with biases, each followed by a ReLU activation function. The size of these layers are (from left to right): $225 \times 50$, $50 \times 50$, $50 \times 25$ and $25 \times 25$. The output of these layers is added to a weighted skip connection layer from the input, and its size is $225 \times 25$. Then, this temporally result is followed by a $25 \times 3$ final layer, obtaining the final derained image $\widehat{\bm{x}}$.

%===================================================================================
%===================================================================================

\section{Image Processing Applications - Results}
\label{sec:Nat_Experiements}

\subsection{Denoising}
Earlier, we proposed two versions of LGM image denoisers (basic and advanced), and for the basic LGM we set $p=8$ and $s=10$. In order to train each one of these models we prepare a training set of clean and noisy image pairs. The training set for basic LGM is the BSD432 dataset \cite{BSD_DATASET}, whereas the  Waterloo Exploration dataset \cite{WATERLOO_DATASET} is combined with BSD432 for the advanced scheme. The inputs are created by adding an additive white Gaussian noise with standard deviation $\sigma$ to the clean images (sampled at each epoch), then we randomly crop the clean and noisy images at the same location, and finally we subtract the mean of the noisy crop from both of them, this way creating the input-output pairs. Crop size for basic LGM is $100 \times 100$ and $56 \times 56$ for the advanced one. For each noise level, LGM models are initialized as explained earlier, and then trained using ADAM optimizer with batch size of $8$ and learning rate of $0.002$. The learning rate is multiplied by a factor of $0.5$ every $20$ epochs when training the advanced LGM. Moreover, the loss function we seek to minimize during the training process is the $log$ $l_2$ loss function, augmented by the mutual coherence $\mu$ of the learned dictionaries as a regularization term to the training loss. Thus, the final training loss is defined as follows:
% \begin{eqnarray}\label{eq:LGM_LOSS}
% \mathcal{L}=log\left ( \sum _{i \in batch} \left \| \bm{x}_i^* - \widehat{\bm{x}}_i \right \|_2^2 \right ) + \xi\left ( \sum_{\bm{D} \in \textit{learned dictionaries}} \mu\left ( \bm{D} \right ) \right ),
% \end{eqnarray}
\begin{eqnarray}\label{eq:LGM_LOSS2}
\mathcal{L}= log \left(\sum_{\left ( \bm{x},\bm{x}^* \right ) \in \textit{training set}} \left \| \bm{x}^* - \widehat{\bm{x}} \right \|_2^2 \right ) + \xi\left ( \sum_{\bm{D} \in \textit{learned dictionaries}} \mu\left ( \bm{D} \right ) \right ),
\end{eqnarray}
in which $\bm{x}^*$ is the clean image crop and $\widehat{\bm{x}}$ is its denoised version (the output of the model). $\xi$ is a parameter which we set it to be $1e^{-5}$. 

Table \ref{tab:Set12_table} presents basic and advanced LGM denoising results on Set12 dataset alongside other known methods, Table \ref{tab:BSD68_table} presents denoising results on BSD68 dataset. Deep-KSVD1, Deep-KSVD2 and DeepKSVD3 are LKSVD$_{1,8,256}$, LKSVD$_{3,8,256}$ and LKSVD$_{2,16,1024}$ respectively, as denoted in \cite{DEEPKSVD}. Note that the better results in \cite{DEEPKSVD} assume a repetition of the denoising in several steps, imitating the EPLL, while our scheme does not use this option. 

As can be seen from these two tables, our advanced scheme is much better than the original KSVD method, and comparable in performance with the better Deep-KSVD results~\cite{DEEPKSVD}. Recall that the main difference between these two schemes is the pursuit applied - LISTA (relaxation) versus LGM (greedy). We believe that the similarity in performance is due to the MMSE flavor of our training approach, which weakens the natural benefits of the greedy alternative in the denoising application. 

\begin{figure}[!htbp]
    \centering
    \begin{subfigure}[b]{0.32\textwidth}
        \captionsetup{justification=centering}
        \includegraphics[width=\textwidth]{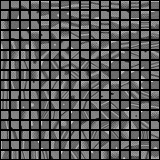}
        \caption{LGM analysis dictionary}
        % \\      $\mu=0.97$}
        \label{fig:LGM_regular_dict}
    \end{subfigure}
    \hfill
    \begin{subfigure}[b]{0.32\textwidth}
        \captionsetup{justification=centering}
        \includegraphics[width=\textwidth]{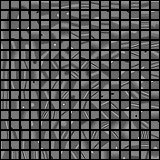}
        \caption{LGM synthesis dictionary}
        % \\        $\mu=0.99$}
        \label{fig:LGM_synthesis_dict}
    \end{subfigure}
    \hfill
    \begin{subfigure}[b]{0.32\textwidth}
        \captionsetup{justification=centering}
        \includegraphics[width=\textwidth]{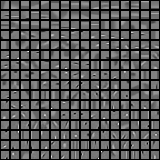}
        \caption{KSVD dictionary}
        %\\        $\mu=0.96$}
        \label{fig:KSVD_dict}
    \end{subfigure}
    \caption{Comparison of the learned dictionaries for noise level $\sigma=25$. %$\mu$ is the mutual coherence.
    }
    \label{fig:dictionary_comparison}
\end{figure}

Figure \ref{fig:dictionary_comparison} presents the learned dictionaries of the basic LGM and the universal KSVD (the version with global dictionary) referring to the same noise level. 
%The mutual coherence of each one of them is also presented in this figure, it should not be a surprise that the mutual coherence of both LGM dictionaries is high, this is because the low coefficient that the mutual coherence regularizer got in the loss term (\ref{eq:LGM_LOSS2}). Moreover, the reason behind KSVD lower mutual coherence is the removal and replacement of the highly correlated atoms from the KSVD dictionary. 
As can be noticed, LGM analysis and synthesis dictionaries are very similar, and they are more edge-friendly than the KSVD's one. A similar figure in \cite{DEEPKSVD} exposes the fact that our learned dictionaries are very different form the ones LISTA leads to. 

\begin{figure}[!htbp]
    \centering
    \begin{subfigure}[b]{0.48\textwidth}
        \includegraphics[width=\textwidth]{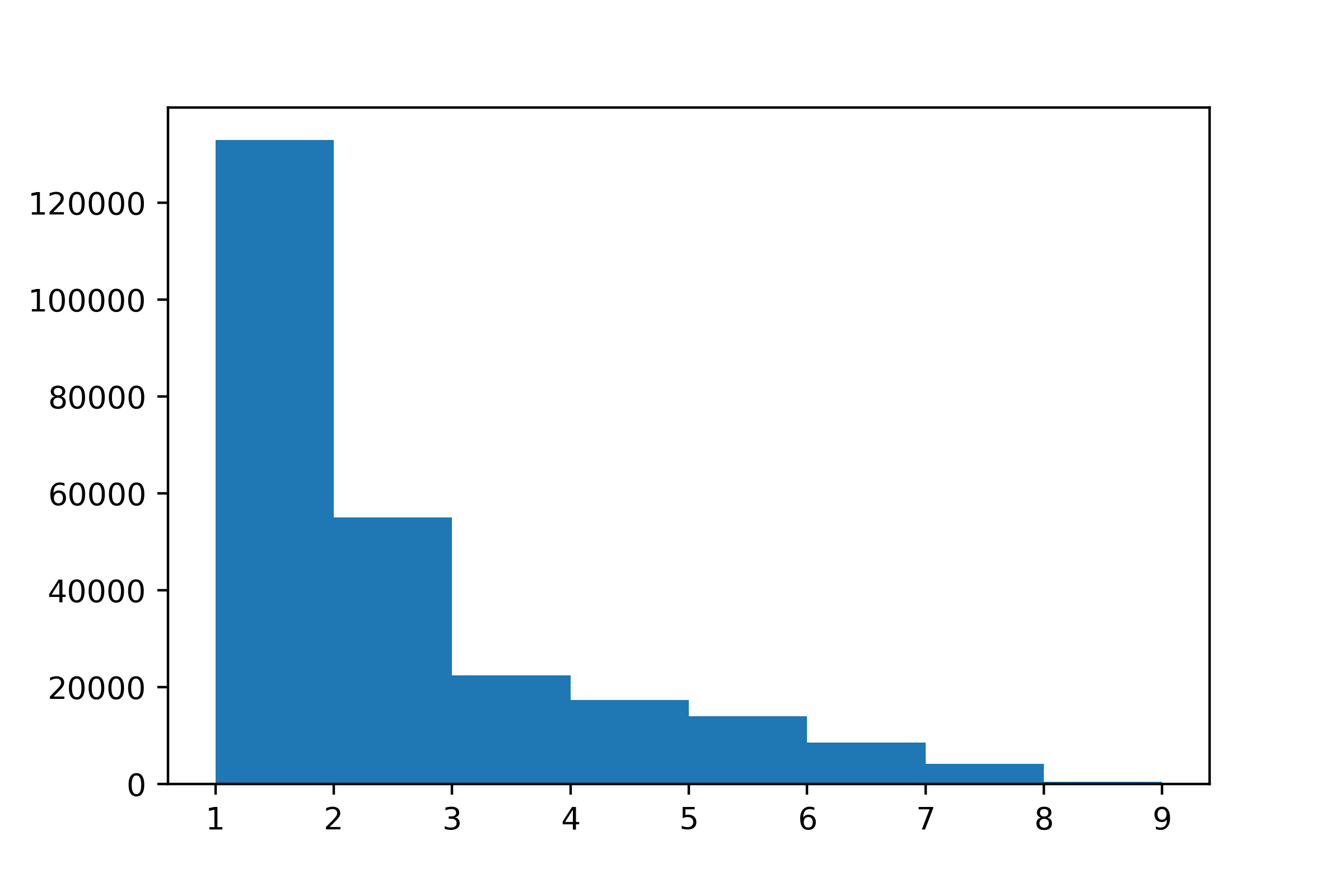}
        \caption{Basic LGM}
        \label{fig:LGM_cardinality_hist}
    \end{subfigure}
    \hfill
    \begin{subfigure}[b]{0.48\textwidth}
        \includegraphics[width=\textwidth]{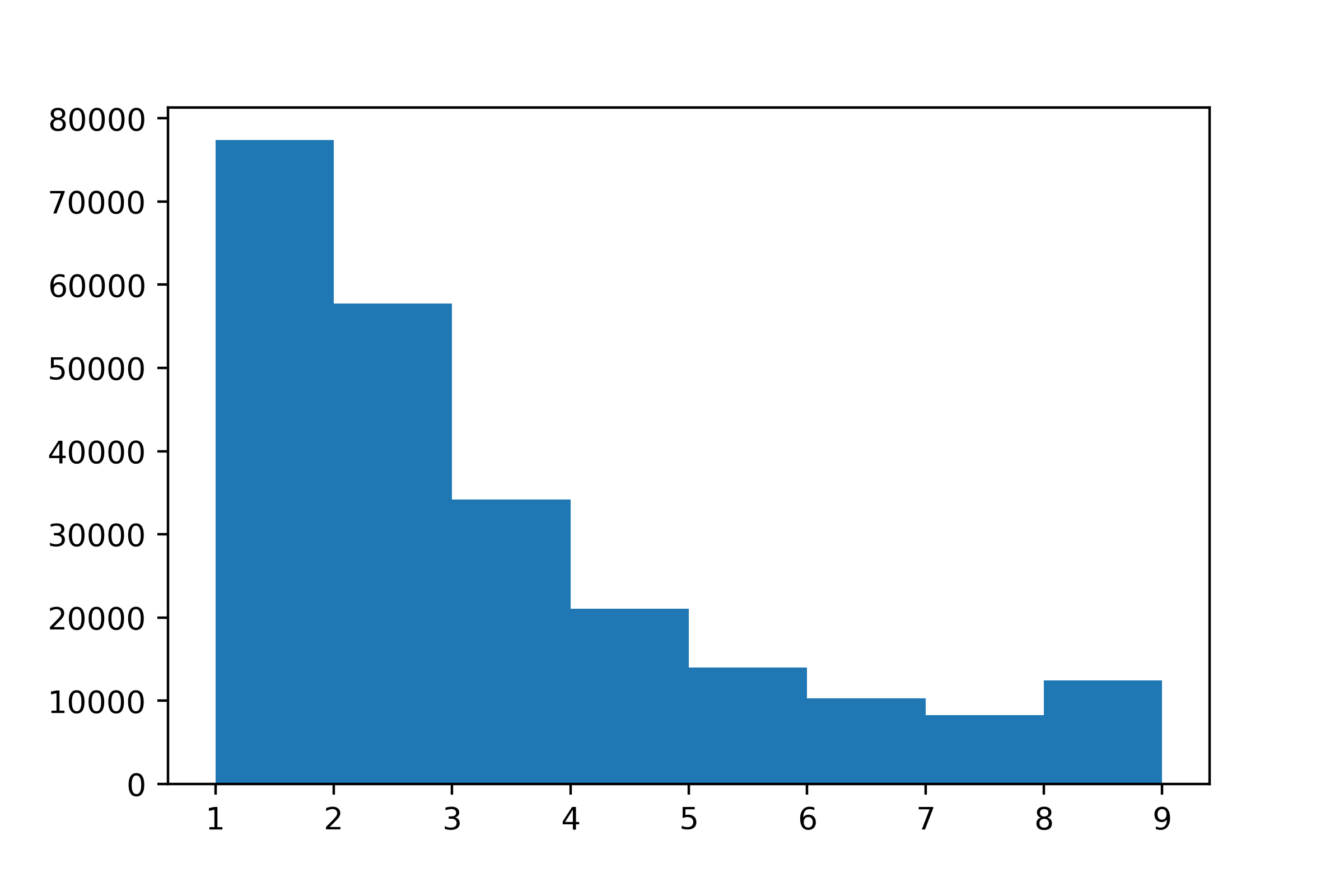}
        \caption{KSVD}
        \label{fig:SKVD_cardinality_hist}
    \end{subfigure}
    \caption{Restored patches cardinality histogram on a specific image}
    \label{fig:cardinality_hist}
\end{figure}

\begin{figure}[!htbp]
    \centering
    \includegraphics[width=1\textwidth]{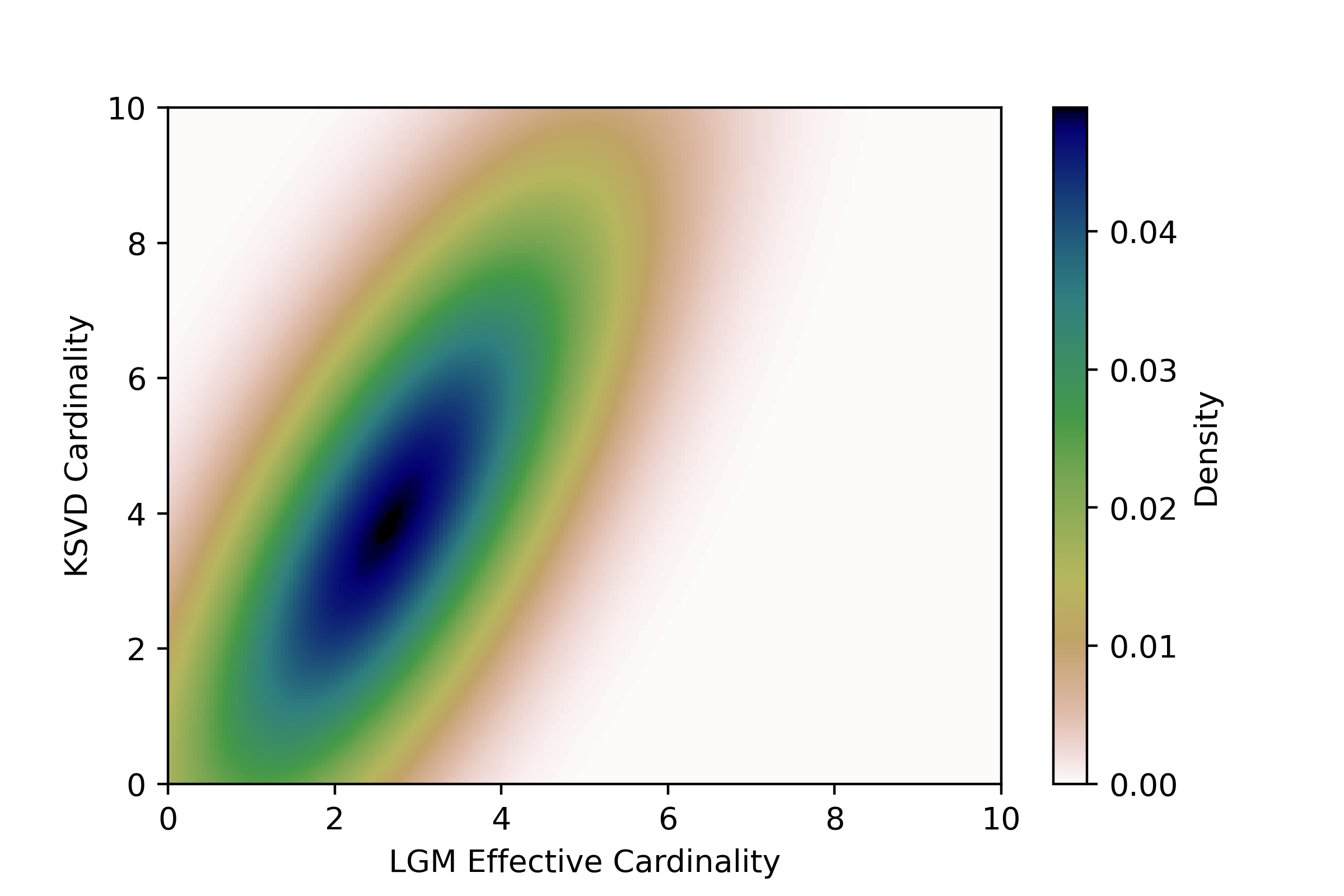}
    \caption{KSVD restored cardinality as a function of LGM effective cardinality of the same patch on a specific image}
    \label{fig:LGM_vs_KSVD_cardinality}
\end{figure}

Figure \ref{fig:cardinality_hist} presents the histograms of LGM and KSVD restored patches cardinality on the same image. The true cardinality of the restored patches of Basic LGM is always $s$, but since the attention net learns the stopping criterion, the effective cardinality of each patch can be calculated by a weighted averaging of the attention weights, $\sum_{i=1}^s i p_i$, and these are the values presented in Figure \ref{fig:LGM_cardinality_hist}. 
% A note worth mentioning: Figure \ref{fig:SKVD_cardinality_hist} shows that there are many patches that get zero cardinality by KSVD, due to the mean removal from each patch, which results in transforming the DC patches into low energy patches, requiring zero atoms fro their representation. This is equivalent to an effective cardinality of one by LGM, choosing the injected atom of ones. 
In the same spirit, Figure \ref{fig:LGM_vs_KSVD_cardinality} presents the representation's cardinality for each patch in the original KSVD versus the cardinality obtained by our Basic LGM representation. As can be seen, there is a rough match between the two, but it is not a perfect alignment. It appears that the LGM scheme tends to a deeper sparsity when compared to the original KSVD, something that could be explained by the better tuned diciotnary we have. Again, we refer the reader to similar graphs in \cite{DEEPKSVD}, where the behavior is totally different, with representations that are not sparse at all. 

\begin{table}[!htbp]
    \begin{subtable}{0.45\textwidth}
      \begin{center}
        \caption{Set12}
        \label{tab:Set12_table}\scriptsize
        \begin{tabular}{||c|c|c|c||} 
        \hline
            $\sigma$ & 15 & 25 & 50 \\
            \hline
            \hline
            $\textit{KSVD}$ & 31.95 & 29.41 & 25.78 \\
            \hline
            $\textit{BM3D}$ & 32.37 & 29.97 & 26.72\\
            \hline
            $\textit{WNNM}$ & 32.70 & 30.26 & 27.05\\
            \hline
            $\textit{TNRD}$ & 32.50 & 30.06 & 26.81\\
            \hline
            $\textit{Deep-KSVD1}$ & - & 29.76 & -\\
            \hline
            $\textit{Deep-KSVD2}$ & 32.53 & 30.12 & 26.91\\
            \hline
            $\textit{Deep-KSVD3}$ & 32.61 & 30.22 & 27.04\\
            \hline
            $\textit{DnCNN}$ & 33.16 & 30.80 & 27.18\\
            \hline
            \hline
            $\textit{Basic LGM (ours)}$ & 32.33 & 29.83 & 26.37\\
            \hline
            $\textit{Advanced LGM (ours)}$ & 32.57 & 30.14 & 26.78\\
            \hline
            \hline

        \end{tabular}
      \end{center}
    \end{subtable}
    \begin{subtable}{0.45\textwidth}
        \begin{center}
        \caption{BSD68}
        \label{tab:BSD68_table}\scriptsize
        \begin{tabular}{||c|c|c|c||} 
        \hline
            $\sigma$ & 15 & 25 & 50\\
            \hline
            \hline
            $\textit{KSVD}$ & 30.91 & 28.32 & 25.03\\
            \hline
            $\textit{BM3D}$ & 31.07 & 28.57 & 25.62\\
            \hline
            $\textit{WNNM}$ & 31.37 & 28.83 & 25.87\\
            \hline
            $\textit{TNRD}$ & 31.42 & 28.92 & 25.97\\
            \hline
            $\textit{Deep-KSVD1}$ & - & 28.76 & -\\
            \hline
            $\textit{Deep-KSVD2}$ & 31.48 & 28.96 & 25.97\\
            \hline
            $\textit{Deep-KSVD3}$ & 31.54 & 29.07 & 26.13\\
            \hline
            $\textit{DnCNN}$ & 31.73 & 29.23 & 26.23\\
            \hline
            \hline
            $\textit{Basic LGM (ours)}$ & 31.30 & 28.76 & 25.67\\
            \hline
            $\textit{Advanced LGM (ours)}$ & 31.47 & 28.96 & 25.94\\
            \hline
            \hline

        \end{tabular}
      \end{center}
    \end{subtable}
    \caption{Denoising results (PSNR)}
    \label{tab:denoising}
\end{table}

\subsection{Deraining}
We move on to the image deraining task. We set $p=8$ and $s=20$ for the LGM deraining scheme presented earlier. The training set is composed of $200$ clean and rainy image pairs. For each clean image, its rainy version is created by adding synthesized rain streaks to it. For further details about the training set, the reader is referred to \cite{Jorder}. During the training procedure, the model's inputs are created by randomly cropping both the clean and the rainy images at the same location, and since we are working with RGB images, the crop size is set to be $52 \times 52$. The model is trained using ADAM optimizer with batch size of $8$ and learning rate of $0.002$, and the learning rate is multiplied by a factor of $0.5$ every $400$ epochs. Moreover, the loss function which we seek to minimize is:
\begin{eqnarray}\label{eq:LGM_LOSS_Derain}
\mathcal{L}= log \left(\sum_{\left ( \bm{x},\bm{x}^* \right ) \in \textit{training set}} \left \| \bm{x}^* - \widehat{\bm{x}} \right \|_2^2 +0.01 \left \| \bm{x}^* - \widehat{\bm{x}}^c \right \|_2^2 +\left \| \bm{x}^* - \left ( \bm{x}-\widehat{\bm{x}}^s \right ) \right \|_2^2   \right )  \\ \nonumber
+\xi\left ( \sum_{\bm{D} \in \textit{learned dictionaries}} \mu\left ( \bm{D} \right ) \right ), \\ \nonumber
\end{eqnarray}
in which $\bm{x}^*$ is the clean image crop, $\widehat{\bm{x}}^c$, $\widehat{\bm{x}}^s$ and $\widehat{\bm{x}}$ are the restored content image, restored rain streaks and derained version respectively. The reason behind the small coefficient for the context term in the loss function is because this yields better results than an equally weighted loss function. $\xi$ is a parameter which we set it to be $1e^{-2}$.

\begin{figure}[!htbp]
    \centering
    \begin{subfigure}[b]{0.49\textwidth}
        \includegraphics[width=\textwidth]{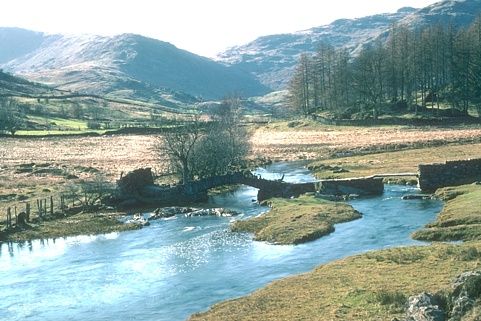}
        \caption{Original clean image}
        \label{fig:Clean}
    \end{subfigure}
    \hfill
    \begin{subfigure}[b]{0.49\textwidth}
        \includegraphics[width=\textwidth]{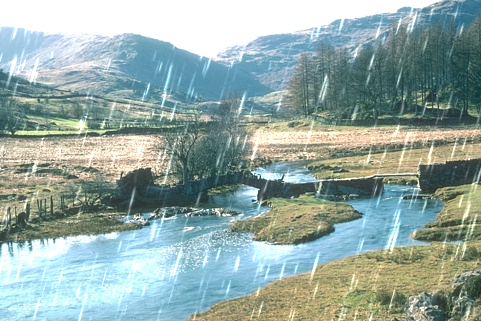}
        \caption{Rainy image (PSNR:25.30)}
        \label{fig:MSE2}
    \end{subfigure}
    
    \begin{subfigure}[b]{0.49\textwidth}
        \includegraphics[width=\textwidth]{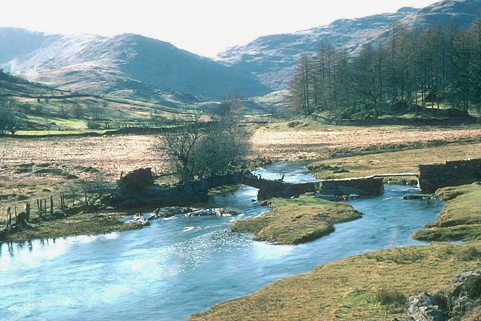}
        \caption{Derained image - $\widehat{\bm{x}}$ (PSNR:33.84)}
        \label{fig:derained}
    \end{subfigure}
    \hfill
    \begin{subfigure}[b]{0.49\textwidth}
        \includegraphics[width=\textwidth]{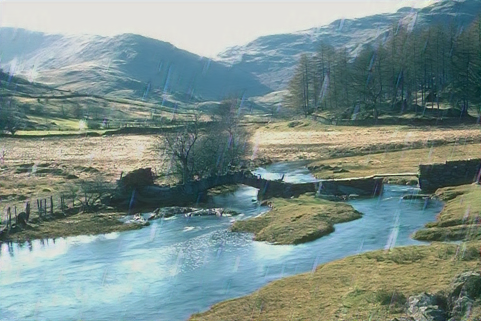}
        \caption{Content part - $\widehat{\bm{x}}^c$ (PSNR:29.38)}
        \label{fig:derained-content}
    \end{subfigure}
    \hfill
    \begin{subfigure}[b]{0.49\textwidth}
        \includegraphics[width=\textwidth]{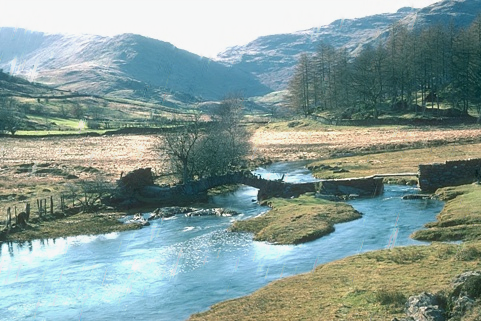}
        \caption{Restored rain streaks subtracted from rainy image - $\widehat{\bm{x}}-
        \widehat{\bm{x}}^s$ (PSNR:33.03)}
        \label{fig:derained-res}
    \end{subfigure}

    \caption{Visual example of LGM different stages}
    \label{fig:Derain}
\end{figure}

\begin{figure}[!htbp]
    \centering
    \includegraphics[width=1\textwidth]{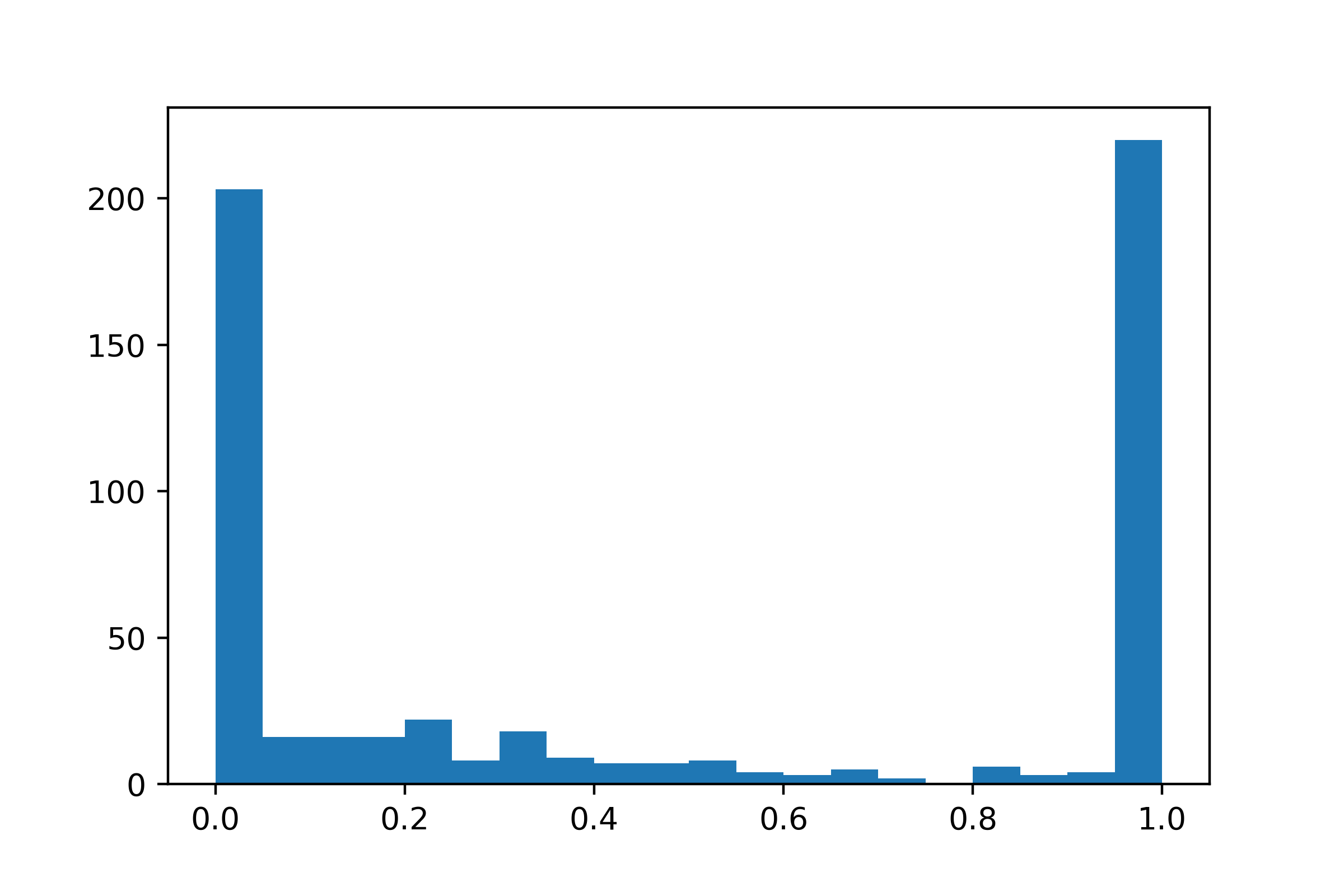}
    \caption{LGM deraining $\textit{f}\left ( \bm{\theta} \right )$ histogram}
    \label{fig:Derain_coefs}
\end{figure}

\begin{figure}[!htbp]
    \centering
    \begin{subfigure}[b]{0.49\textwidth}
        \includegraphics[width=\textwidth]{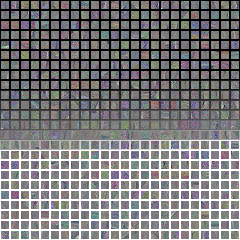}
        \caption{Regular dictionary}
        \label{fig:LGM_regular_dict_deraining}
    \end{subfigure}
    \hfill
    \begin{subfigure}[b]{0.49\textwidth}
        \includegraphics[width=\textwidth]{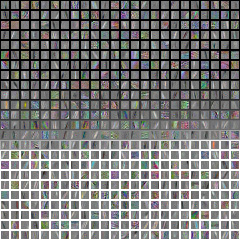}
        \caption{Synthesis dictionary}
        \label{fig:LGM_synthesis_dict_deraining}
    \end{subfigure}
    \caption{LGM learned dictionaries, the color around each atom represents the value of its $\textit{f}\left ( {\theta} \right )$, in which black and white corresponds to $0$ (rain) and $1$ (content) respectively. Each value between $0$ and $1$ gets its corresponding gray-scale intensity}
    \label{fig:learned_dict_deraining}
\end{figure}

\begin{figure}[!htbp]

    \centering
    \begin{subfigure}[b]{0.24\textwidth}
        \includegraphics[width=\textwidth]{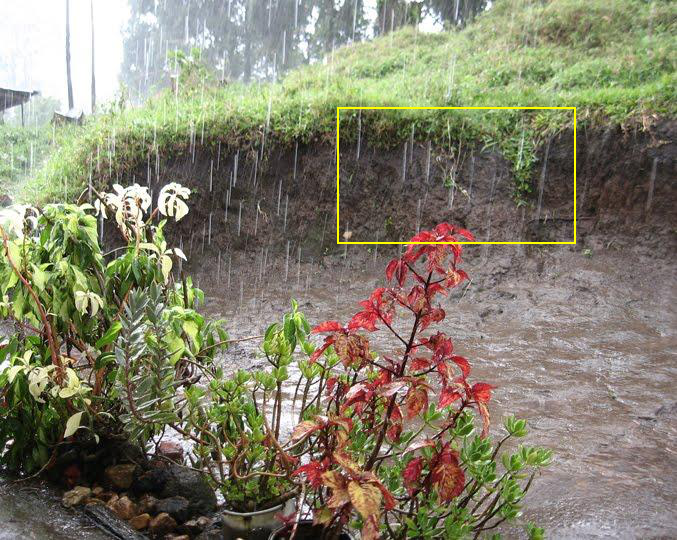}
        \includegraphics[width=\textwidth]{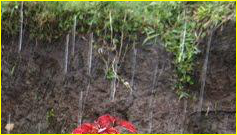}
     \includegraphics[width=\textwidth]{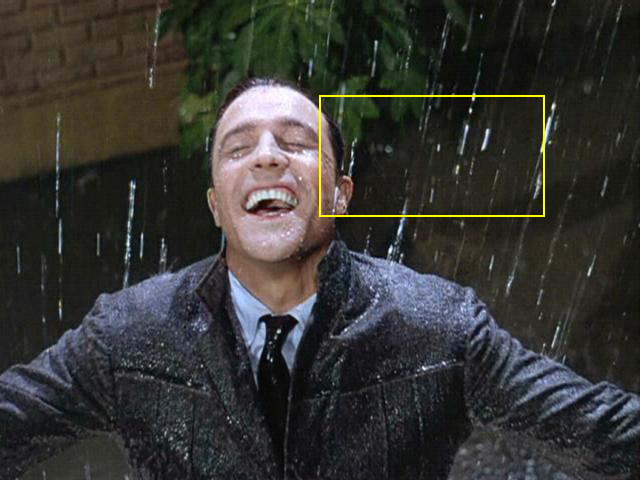}
     
     \includegraphics[width=\textwidth]{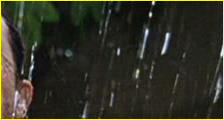}

        \caption{Rainy images}
        \label{fig:Real_Rainy}
    \end{subfigure}
    \begin{subfigure}[b]{0.24\textwidth}
        \includegraphics[width=\textwidth]{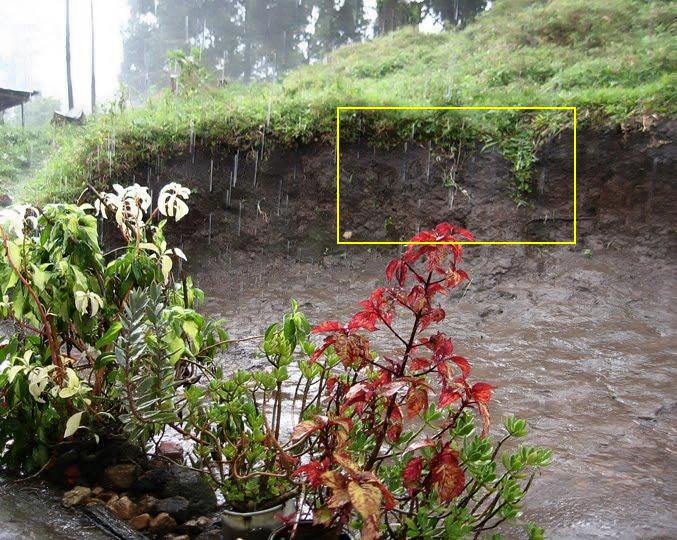}
        \includegraphics[width=\textwidth]{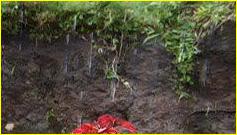}
     \includegraphics[width=\textwidth]{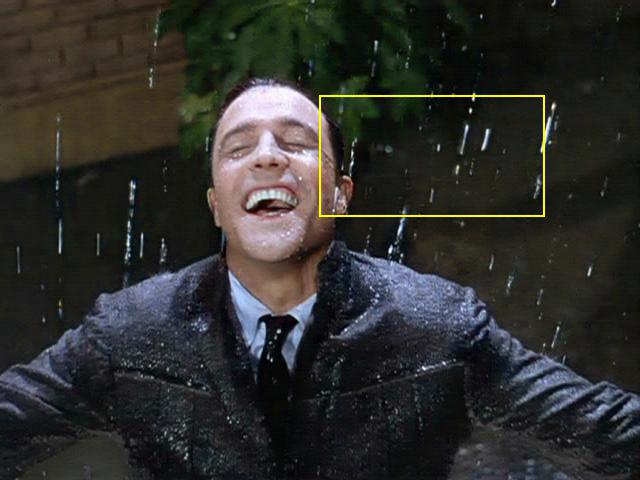}
      \includegraphics[width=\textwidth]{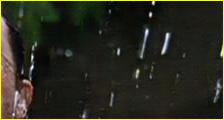}

        \caption{DNN \cite{Derain4}}
        \label{fig:derained-DNN}
    \end{subfigure}
    \begin{subfigure}[b]{0.24\textwidth}
        \includegraphics[width=\textwidth]{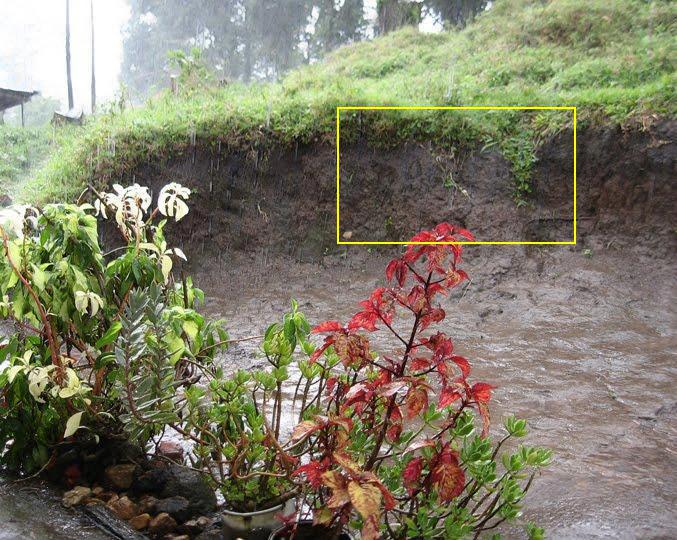}
        \includegraphics[width=\textwidth]{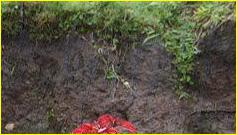}

     \includegraphics[width=\textwidth]{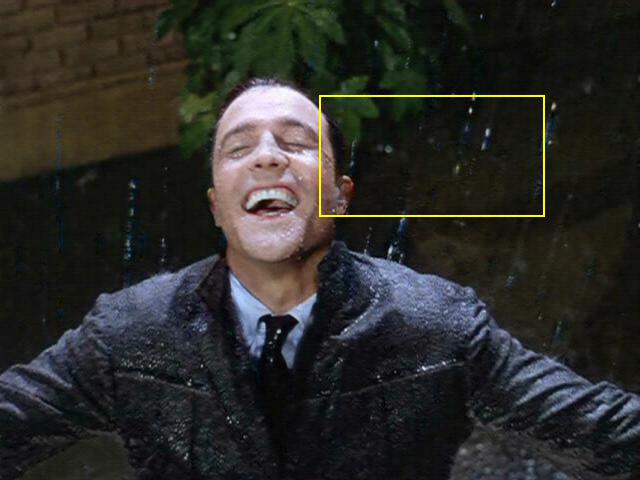}
     \includegraphics[width=\textwidth]{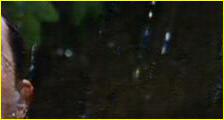}

        \caption{Jorder \cite{Jorder}}
        \label{fig:derained-Jorder}
    \end{subfigure}
    \begin{subfigure}[b]{0.24\textwidth}
        \includegraphics[width=\textwidth]{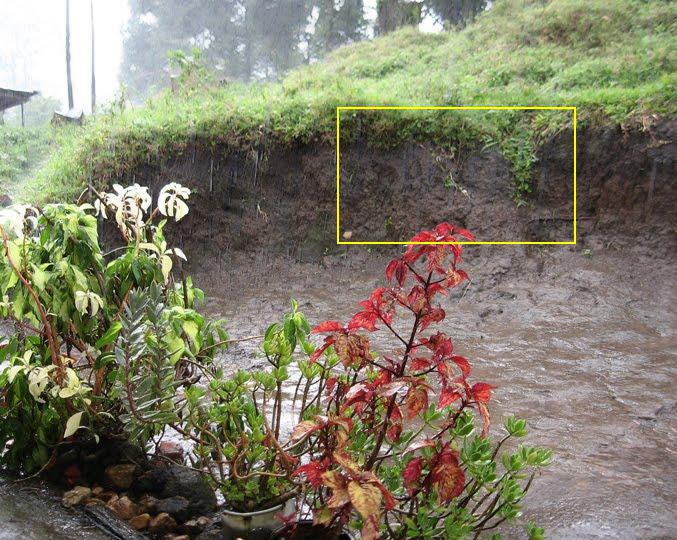}
        \includegraphics[width=\textwidth]{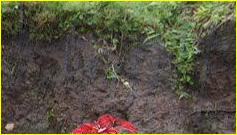}
      \includegraphics[width=\textwidth]{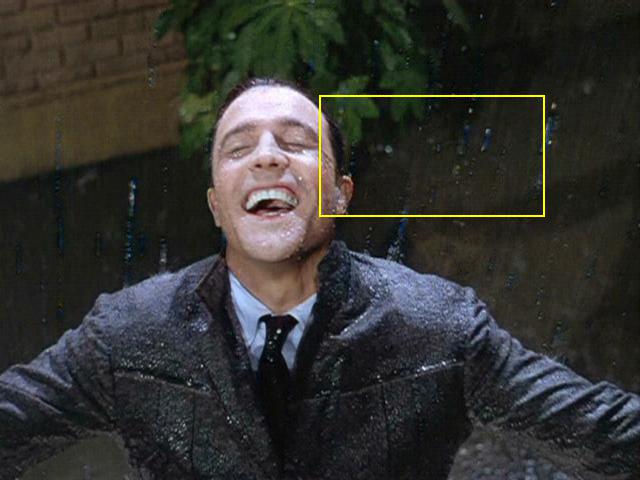}
     \includegraphics[width=\textwidth]{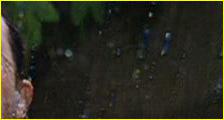}

        \caption{LGM (ours)}
        \label{fig:derained-LGM}
    \end{subfigure}
    
    \caption{Visual comparison of real rainy images deraining}
    \label{fig:Derain_RealRain}

\end{figure}

We use two test sets -- Rain12 \cite{Derain3} and Rain100L \cite{Jorder} in order to evaluate the proposed model and compare it with other methods. The results are given in Tables \ref{tab:Rain12_table} and \ref{tab:Rain100L_table}. We report two result versions of our proposed LGM model, the first calculates the error in the RGB domain, while in the second the metric is calculated after transforming the image into the luma component in the YCbCr domain. The second metric (luma component) is the metric used by \cite{Jorder,Prenet}, and we calculated it on our model's output using the software provided by \cite{Prenet}. As can be noticed, our approach outperforms the classical methods and some of the deep learning approaches. Figure \ref{fig:Derain} includes a visual example of the LGM deraining scheme applied on an image from Rain100, where the difference between $\widehat{\bm{x}}^c$, $\bm{x}- \widehat{\bm{x}}^s$ and $\bm{x}$ can be noticed. Moreover, the histogram of $\textit{f}\left ( \bm{\theta} \right )$ (which describes the image content percentage of each atom) is attached in Figure \ref{fig:Derain_coefs}. As can be seen, most of the coefficient's values are larger than $0.8$ or less than $0.2$, indicating that the desired separation is achieved. Figure \ref{fig:learned_dict_deraining} presents the learned dictionaries by LGM and the corresponding content-rain atoms separation. We also test the model on real images and the results are presented in Figure \ref{fig:Derain_RealRain}. As can be noticed in this figure, LGM performance on real rainy images is very similar to Jorder \cite{Jorder} and somewhat better.

\begin{table}[h]
    \begin{subtable}{0.45\textwidth}
      \begin{center}
        \caption{Rain12}
        \label{tab:Rain12_table}\scriptsize
        \begin{tabular}{||c|c||} 
        \hline
            
            $\textit{LP}$ \cite{Derain3} - results reported in \cite{Jorder} & 32.02/0.91 \\
            \hline
            $\textit{DDN}$ \cite{Derain4} - results reported in \cite{Prenet} &  31.78/0.90 \\
            \hline
            $\textit{DSC}$ \cite{Derain2} - results reported in \cite{Jorder} & 30.02/0.87 \\
            \hline
            $\textit{JORDER}$ \cite{Jorder} & 36.02/0.96 \\
            \hline
            $\textit{JORDER}$ - results reported in \cite{Prenet} & 33.92/0.95 \\
            \hline
            $\textit{PEeNet}$ \cite{Prenet} & 36.69/0.96 \\
            \hline
            \hline
            $\textit{LGM (ours)}$ & 34.11/0.94 \\
            \hline
            $\textit{LGM (ours)}$ - using metric in \cite{Prenet} & 35.46/0.95 \\
            \hline

        \end{tabular}
      \end{center}
  \end{subtable}
  \begin{subtable}{0.45\textwidth}
      \begin{center}
        \caption{Rain100L}
        \label{tab:Rain100L_table}\scriptsize
        \begin{tabular}{||c|c||} 
        \hline
            
            $\textit{LP}$ \cite{Derain3} - results reported in \cite{Jorder} & 29.11/0.88 \\
            \hline
            $\textit{DDN}$ \cite{Derain4} - results reported in \cite{Prenet}&   32.16/0.94 \\
            \hline
            $\textit{DSC}$ \cite{Derain2} - results reported in \cite{Jorder} & 24.16/0.87 \\
            \hline
            $\textit{JORDER}$ \cite{Jorder} & 36.112/0.97 \\
            \hline
            $\textit{JORDER}$ - results reported in \cite{Prenet} &  36.61/0.97 \\
            \hline
            $\textit{PEeNet}$ \cite{Prenet} & 37.10/0.98 \\
            \hline
            \hline
            $\textit{LGM (ours)}$ & 32.65/0.95 \\
            \hline
            $\textit{LGM (ours)}$ - using metric in \cite{Prenet} & 34.07/0.96 \\
            
            \hline

        \end{tabular}
      \end{center}
  \end{subtable}
  \caption{Deraining results (PSNR/SSIM)}
  \label{tab:deraining}
\end{table}

% \begin{table}[h]
  
% \end{table}

%===================================================================================
%===================================================================================

\section{Conclusions}
\label{sec:Conclusion}
In this work we introduced a technique of unfolding greedy sparse pursuit algorithms into a deep neural network. Our main goal is to tackle the problem of interpretability which deep learning field still suffers from until now. Continuously to the series of works in which classical algorithms are unfolded to deep neural networks, this work introduces an architecture with well justified features, such as dynamic number of layers and an activation function with greedy nature. To our knowledge, this is the first kind of work that gives a clear justification to such features. Also, this opens the door for further works in which classical methods with combinatorial nature turns into a Neural Network, especially in  fields such as Computer Vision, Classification, NLP and more. Moreover, we hope to see future works in which LGM is deployed in order to solve problems such as deblurring, super-resolution and compression. However, the obtained LGM performance in both denoising and deraining does not compete with the state-of-the-art methods, which gives the rise to the question, whether it is time to wonder that sparse model for natural images is outdated?.

%===================================================================================
%===================================================================================

\section{Supplementary Materials}
\subsection{Dictionary Distance Metric}
\label{sub_section:Dict_Dist}

The dictionary distance metric that is used throughout this paper is described in algorithm \ref{algo:DDist_Metric}. Given the true dictionary $\bm{D}_{true}$ and a learned dictionary $\bm{D}_{approx}$ (both in their normalized versions), the distance is calculated by sweeping over all the atoms in $\bm{D}_{true}$ and for each we find the ``closest'' atom from $\bm{D}_{approx}$ and evaluate its corresponding distance from it. Finally, the metric value is obtained by averaging these distances.

\begin{algorithm}[h]
\SetAlgoLined
\SetKwFunction{FName}{DictionaryDist}
\SetKwProg{Pn}{Function}{:}{}

  \Pn{\FName{$\bm{D}_{true}=\left [
 \bm{d}^{true}_1, 
 \bm{d}^{true}_2, 
\dots, 
 \bm{d}^{true}_{m_1}
\right ]\in \mathbb{R}^{n \times m_1}$, $\bm{D}_{approx}=\left [
 \bm{d}^{approx}_1, 
 \bm{d}^{approx}_2, 
\dots, 
 \bm{d}^{approx}_{m_2}
\right ]\in \mathbb{R}^{n \times m_2}$}}{
  \BlankLine
%   $\textit{normlizing }$
  
  $dist = 0$
  
  \For{$i=1,2,...,m_1$}{
    $dist += \min \left ( 1- \left | \bm{D}_{approx}^T \bm{d}^{true}_i \right | \right ) $
}
\BlankLine
  }
  $dist /= m_1$
  
  {\KwRet  $dist$}
 
 \caption{Dictionary Distance Metric}
 \label{algo:DDist_Metric}
\end{algorithm}

\bibliographystyle{unsrt}  
\bibliography{references}

\end{document}